\documentclass[10pt,twocolumn,letterpaper]{article}

\usepackage[pagenumbers]{cvpr} 

\usepackage[accsupp]{axessibility}
\usepackage[dvipsnames,table]{xcolor}
\usepackage{microtype}

\usepackage{tikz}
\usetikzlibrary{calc}
\usetikzlibrary{positioning}
\usetikzlibrary{angles,quotes}
\usetikzlibrary{backgrounds}
\usetikzlibrary{external}
\usetikzlibrary{fit}
\usetikzlibrary{arrows}
\usetikzlibrary{arrows.meta}
\usetikzlibrary{shapes.symbols}
\usetikzlibrary{shadings}
\usetikzlibrary{shapes}
\usetikzlibrary{spy}
\usetikzlibrary{fadings}
\usetikzlibrary{matrix}
\usetikzlibrary{decorations.pathmorphing, decorations.pathreplacing, patterns}
\usetikzlibrary{3d}

\usepackage{pgfplots}
\usepackage{pgfplotstable}
\pgfplotsset{compat=1.16}
\pgfplotsset{every axis plot/.append style={thick,mark size=1.5pt,line join=bevel,mark options={solid}}}
\pgfplotsset{label style={font=\small}}
\pgfplotsset{tick label style={font=\footnotesize}}
\pgfplotsset{grid style={color=black!10}}
\pgfplotsset{every non boxed x axis/.style={xtick align=center,shorten <=-.5\pgflinewidth}}
\pgfplotsset{every non boxed y axis/.style={ytick align=center,shorten <=-.5\pgflinewidth}}
\pgfplotsset{every non boxed z axis/.style={ztick align=center,shorten <=-.5\pgflinewidth}}
\pgfplotsset{/pgf/number format/1000 sep={\,}}
\usepgfplotslibrary{groupplots}

\usepackage{epsdice}
\usepackage{sansmath}

\usepackage[normalem]{ulem}

\usepackage{pdflscape}
\usepackage{listings}
\usepackage{multirow}

\usepackage{pifont}
\usepackage{fontawesome5}
\usepackage{booktabs}
\usepackage{colortbl}

\usepackage{array}
    \newcolumntype{C}{>{$}c<{$}}
    \newcolumntype{L}{>{$}l<{$}}
    \newcolumntype{R}{>{$}r<{$}}
    \newcolumntype{M}{>{\centering\arraybackslash$}m{1.4cm}<{$}}

\newcommand{\vz}{\ensuremath{\mathbf{z}}\xspace}

\newcommand{\synthnet}{\ensuremath{f_\theta}\xspace}
\newcommand{\entropynet}{\ensuremath{g_\psi}\xspace}

\newcommand{\cc}{COOL-CHIC\xspace}

\newcommand{\method}{C3\xspace}

\usepackage{siunitx}
\usepackage{csquotes}

\definecolor{C0}{HTML}{1F77B4}
\definecolor{C1}{HTML}{FF7F0E}
\definecolor{C2}{HTML}{2CA02C}
\definecolor{C3}{HTML}{D62728}
\definecolor{C4}{HTML}{9467BD}
\definecolor{C5}{HTML}{8C564B}
\definecolor{C6}{HTML}{E377C2}
\definecolor{C7}{HTML}{7f7f7f}
\definecolor{C8}{HTML}{bcbd22}
\definecolor{C9}{HTML}{17becf}
\colorlet{C0light}{C0!70!white}
\colorlet{C1light}{C1!70!white}
\colorlet{C2light}{C2!70!white}
\colorlet{C3light}{C3!70!white}
\colorlet{C4light}{C4!70!white}
\colorlet{C5light}{C5!70!white}
\colorlet{C6light}{C6!70!white}
\colorlet{C0vlight}{C0!20!white}
\colorlet{C1vlight}{C1!20!white}
\colorlet{C2vlight}{C2!20!white}
\colorlet{C3vlight}{C3!20!white}
\colorlet{C4vlight}{C4!20!white}
\colorlet{C0dark}{C0!70!black}
\colorlet{C1dark}{C1!70!black}
\colorlet{C2dark}{C2!70!black}
\colorlet{C3dark}{C3!70!black}
\colorlet{C4dark}{C4!70!black}
\colorlet{C5dark}{C5!70!black}
\colorlet{C6dark}{C6!70!black}
\colorlet{darkgray}{gray!60!black}
\colorlet{verylightgray}{lightgray!40!white}

\makeatletter
\def\mathcolor#1#{\@mathcolor{#1}}
\def\@mathcolor#1#2#3{%
  \protect\leavevmode
  \begingroup
    \color#1{#2}#3%
  \endgroup
}
\makeatother

\allowdisplaybreaks[2]
\definecolor{cvprblue}{rgb}{0.21,0.49,0.74}
\usepackage[pagebackref,breaklinks,colorlinks,allcolors=cvprblue]{hyperref}

\def\paperID{15981} 
\def\confName{CVPR}
\def\confYear{2025}

\title{Good, Cheap, and Fast: \\ Overfitted Image Compression with Wasserstein Distortion}

\author{Jona Ball\'e\footnotemark[1]~, Luca Versari\footnotemark[2]~, Emilien Dupont\footnotemark[3]~, Hyunjik Kim\footnotemark[3]~, Matthias Bauer\footnotemark[3] \\
\footnotemark[1]~~New York University \footnotemark[2]~~Google Paradigms Of Intelligence \footnotemark[3]~~Google DeepMind \\
{\tt\small jona.balle@nyu.edu, \{veluca,edupont,hyunjikk,msbauer\}@google.com}
}

\begin{document}
\maketitle
\begin{abstract}
Inspired by the success of generative image models, recent work on learned image compression increasingly focuses on better probabilistic models of the natural image distribution, leading to excellent image quality. This, however, comes at the expense of a computational complexity that is several orders of magnitude higher than today's commercial codecs, and thus prohibitive for most practical applications. With this paper, we demonstrate that by focusing on modeling visual perception rather than the data distribution, we can achieve a very good trade-off between visual quality and bit rate similar to \enquote{generative} compression models such as HiFiC, while requiring less than 1\% of the multiply--accumulate operations (MACs) for decompression. We do this by optimizing C3, an overfitted image codec, for Wasserstein Distortion (WD), and evaluating the image reconstructions with a human rater study, showing that WD clearly outperforms LPIPS as an optimization objective. The study also reveals that WD outperforms other perceptual metrics such as LPIPS, DISTS, and MS-SSIM as a predictor of human ratings, remarkably achieving over 94\% Pearson correlation with Elo scores.
\end{abstract}

\section{Introduction}
\label{sec:intro}

\begin{figure*}
\centering
\begin{tikzpicture}[
  spy using outlines={rectangle, green, magnification=3, size=2.5cm, connect spies},
  every node/.style={font=\small},
]
\node[inner sep=0pt] (original)
  {\includegraphics[width=.495\linewidth]{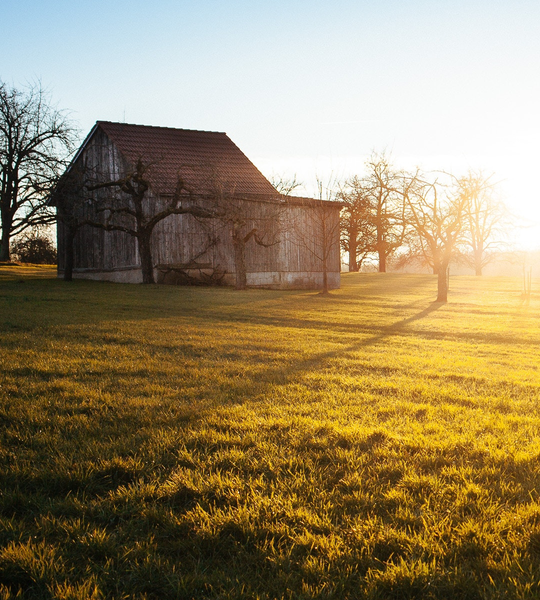}};
\node[inner sep=0pt, right=2pt of original] (mse)
  {\includegraphics[width=.495\linewidth]{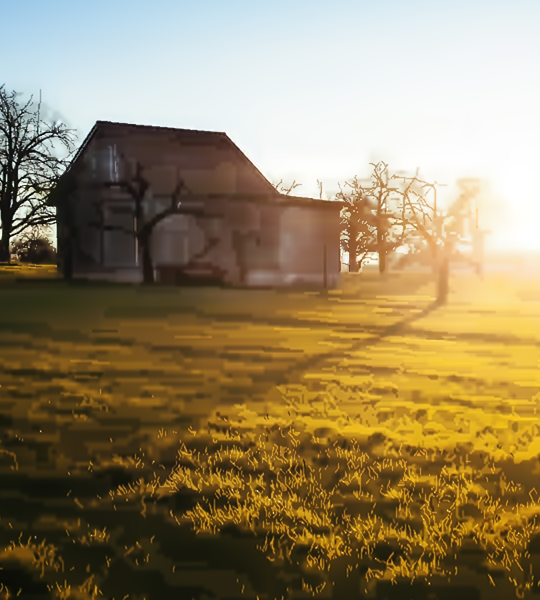}};
\node[inner sep=0pt, below=2pt of original] (wd)
  {\includegraphics[width=.495\linewidth]{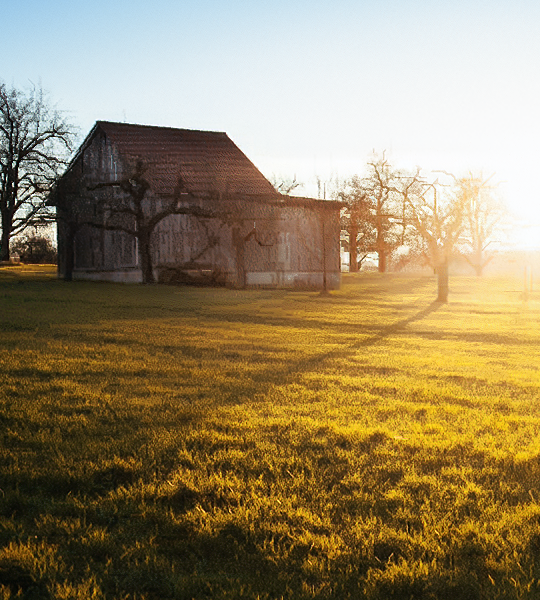}};
\node[inner sep=0pt, below=2pt of mse] (wd_nocr)
  {\includegraphics[width=.495\linewidth]{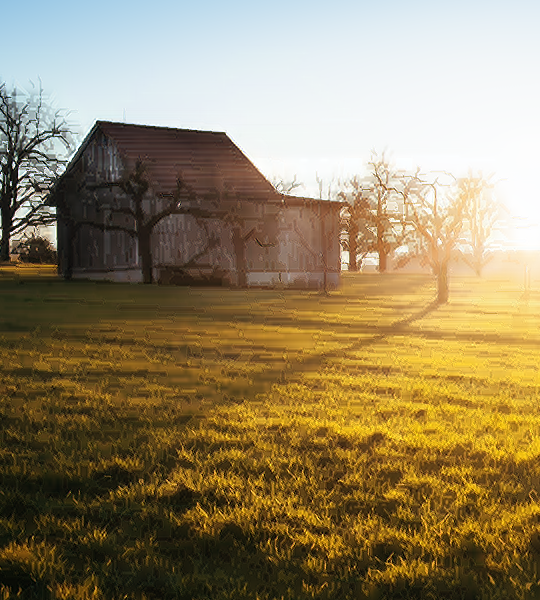}};

\node[below right=0 of original.north west] {\sffamily original};
\node[below right=0 of mse.north west] {\sffamily C3/MSE~\citep{C3_2024} at 0.172\,bits/pixel};
\node[below right=0 of wd.north west] {\sffamily C3/WD8 at 0.167\,bits/pixel};
\node[below right=0 of wd_nocr.north west] {\sffamily C3/WD8 (no CR) at 0.180\,bits/pixel};

\node (original_spy1) at ({$(original.west)!.3!(original.east)$} |- {$(original.south)!.27!(original.north)$}) {};
\spy on (original_spy1) in node[below left=1pt] at (original.north east);
\node (mse_spy1) at ({$(mse.west)!.3!(mse.east)$} |- {$(mse.south)!.27!(mse.north)$}) {};
\spy on (mse_spy1) in node[below left=1pt] at (mse.north east);
\node (wd_spy1) at ({$(wd.west)!.3!(wd.east)$} |- {$(wd.south)!.27!(wd.north)$}) {};
\spy on (wd_spy1) in node[below left=1pt] at (wd.north east);
\node (wd_nocr_spy1) at ({$(wd_nocr.west)!.3!(wd_nocr.east)$} |- {$(wd_nocr.south)!.27!(wd_nocr.north)$}) {};
\spy on (wd_nocr_spy1) in node[below left=1pt] at (wd_nocr.north east);
\end{tikzpicture}
\caption{Crop of image 28 from the CLIC2020 professional dataset (best viewed on screen). \textbf{Top left:} Original image. \textbf{Top right:} C3 optimized for MSE, compressed to 0.172 bits/pixel.
\textbf{Bottom left:} C3 optimized for WD with $\sigma=8$, compressed to 0.167 bits/pixel. \textbf{Bottom right:} C3 optimized for WD with $\sigma=8$, without common randomness, compressed to 0.180 bits/pixel. While optimization for MSE leads to flattened texture, as seen in the reproduction of the grass and the walls of the building, texture is vastly improved in the WD-optimized versions. Not providing common randomness to the decoder (bottom right) means the codec must reproduce all textures using deterministic structure such as straight lines (as in, e.g., the texture of the roof -- note the lines not being consistent with the original, while providing a better approximation of the texture than the MSE version). This is not adequate for random textures such as the grass in front of the building, where providing common randomness significantly helps to maintain the visual quality of texture.}
\label{fig:cr_ablation}
\end{figure*}

\begin{figure*}[t]
\centering
\includegraphics[width=\textwidth]{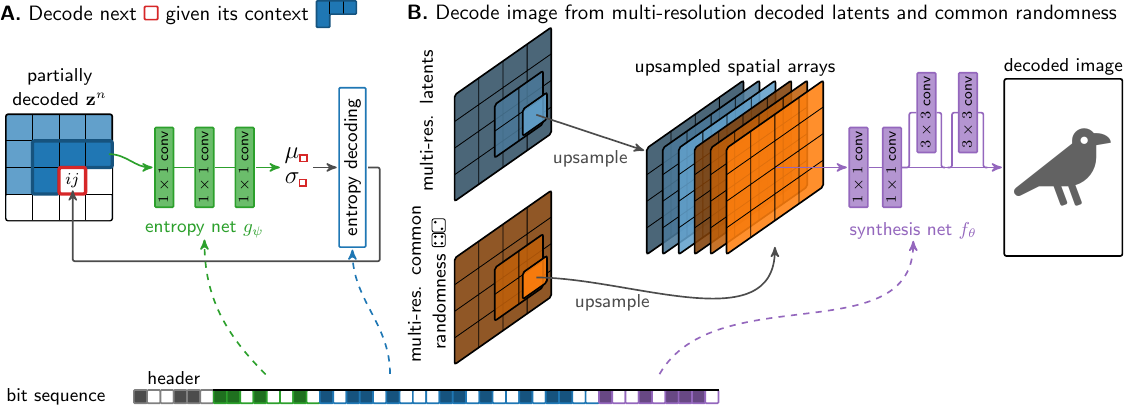}
\caption{Decoding an image with \cc~\citep{ladune2023cool} and \method~\citep{C3_2024}. \textbf{\sffamily A.}~A latent element $\widehat\vz^n_{ij}$ (\protect\tikz[scale=0.2, baseline=0.1ex]{\protect\draw[step=1,draw=C3,fill=white, thick, rounded corners=0.5] (0, 0) rectangle (1, 1);}) is autoregressively decoded by applying the entropy network $\entropynet$ to the spatial context $\mathrm{c}(\vz^n; (i, j))$
(\protect\tikz[scale=0.175, baseline=.75ex]{
  \protect\draw[step=1,fill=C0, draw=none] (1, 1) rectangle (4, 2);
  \protect\draw[step=1,fill=C0] (1, 1) grid (4, 2);
  \protect\draw[step=1,fill=C0, draw=none] (1, 0) grid (2, 1) rectangle (1, 0);
  \protect\draw[C0dark, thick, rounded corners=0.5] (1, 0) -- (2, 0) -- (2, 1) -- (4, 1) -- (4, 2) -- (1, 2) -- cycle;
}),
yielding parameters $\mu$, $\sigma$ of the Laplacian probability model used for entropy decoding the latent element.
\textbf{\sffamily B.}~The decoded latent spatial arrays at multiple resolutions are first bilinearly upsampled to the target resolution and then transformed into pixel space using the synthesis network $\synthnet$. We supply common randomness at multiple resolutions by drawing i.i.d. elements from a pseudo-random number generator with fixed seed \epsdice{4}\epsdice{2}, which is novel to the present work. The common randomness arrays are upsampled and concatenated with the upsampled latents. Figure adapted with permission from \citet{C3_2024}.}
\label{fig:c3_model_overview}
\end{figure*}

Methods of lossy data compression are characterized by a long tradition of exploiting both:
\begin{itemize}
    \item correlations in the statistics of the compressed data, for example by linearly or nonlinearly predicting the next sample in audio codecs, or by applying decorrelating transformations such as the DCT~\citep{ahmed-1974-discrete-cosine}; and
    \item information bottlenecks of human perception, for example by representing chrominance in images with lower resolution than luminance, or by modeling frequency masking in the auditory system.
\end{itemize}

Owing to the conceptual connection between \emph{decoding} of an image and \emph{sampling} one from a conditional distribution, research on learned image compression has recently been greatly influenced by \emph{generative} methods such as GANs~\citep{gans} or diffusion models~\citep{diffusion}, which excel at probabilistic modeling of real-world data, as evidenced by the impressive quality of samples drawn from such models. The generated samples are visually so convincing that measures of sample quality (i.e., \emph{realism}, formalized by divergences between the distribution of the source data and the distribution of generated data) have even been called \enquote{perception}~\citep{blau-2018-the-perception--distortion-tradeoff}.

Unfortunately, a model which is extremely good at sampling real-world data has to have a certain computational complexity, because the world (and hence the space of possible images) is very complex. This is evidenced by the complexity of generative models such as Imagen, Stable Diffusion, or DALL-E. When it comes to image compression, this would make it seem that we ultimately need highly complex methods to achieve high image quality. In other words: out of the three qualities, \emph{good} (image quality), \emph{fast} (decoding speed), and \emph{cheap} (low bit rate), we must \emph{pick two}, as in the old proverb. Luckily, while good probabilistic models can lead to remarkable perceived quality, the inverse does not hold: good image quality \emph{does not} require precisely modeling the distribution of natural images, as we demonstrate in this paper.

Here, we focus on modeling human perception (specifically, texture perception in the periphery of the human visual system~\citep{freeman-2011-metamers-of-the-ventral,balas-2009-a-summary-statistic-representation}) instead of the data distribution. We take an existing low-complexity neural codec, C3~\citep{C3_2024}, and replace its distortion loss by either of two variants of Wasserstein Distortion (WD)~\citep{qiu-2024-wasserstein-distortion}: one which assumes the observer to gaze at each image location with equal probability, and one which uses EML-net~\citep{jia2020eml}, a saliency model, to predict image locations that are more likely to be scrutinized. To aid texture reproduction, we supply the codec with \emph{common randomness} (CR), i.e., randomness available to both the encoder and the decoder, implemented using a pseudo-random number generator with identical seeds.

These changes lead to significantly better image quality at the same bit rate compared to baseline C3~\citep{C3_2024}, illustrated in \cref{fig:cr_ablation}, but retain its low decoding complexity. The encoding complexity increases with the complexity of the loss function, which we mitigate by introducing a novel, approximate implementation of WD. We perform a human rating study comparing our method to several baselines, including generative and commercial codecs. It reveals that our method achieves a trade-off between image quality and bit rate similar to generative methods, but with much lower decoding complexity. In addition, we find that WD predicts the human ratings remarkably well.

\section{C3 codec}

\begin{figure*}[t]
\newcommand{\crowexp}{\ensuremath{\textcolor{black}{\text{\tiny\faCrow}}\textcolor{C3}{\text{\tiny\faCrow}}}}
\includegraphics[width=\textwidth]{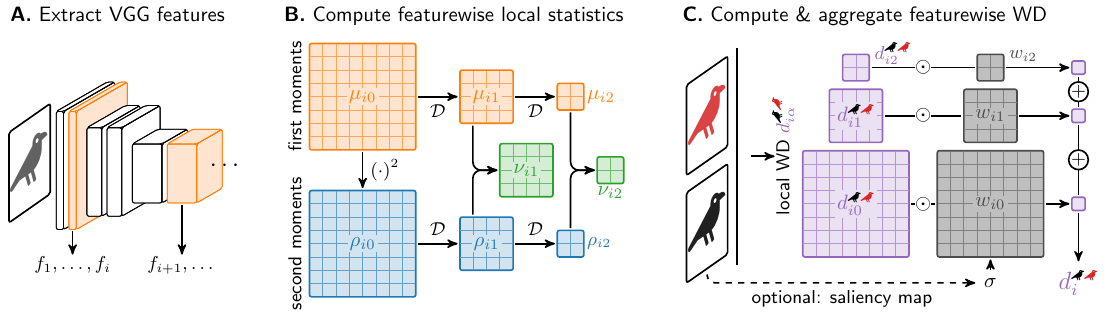}
\caption{Computation of Wasserstein Distortion (WD) between two images, \faCrow{} and \textcolor{C3}{\faCrow}. 
\textbf{\sffamily A.} We extract spatial feature maps $f_i$ from selected layers of a VGG network. 
\textbf{\sffamily B.} For each feature $i$, we estimate local first and second raw moments at multiple scales~$\alpha$ by successively applying a linear downsampling operation $\mathcal{D}$: $\mathcolor{C1}{\mu_{i\alpha}} =\mathcal{D}^\alpha\! f_i$ and $\mathcolor{C0}{\rho_{i\alpha}} =\mathcal{D}^\alpha\! f_i^2$; the local standard deviations are derived by taking $\mathcolor{C2}{\nu_{i\alpha}} = \sqrt{\smash[b]{\mathcolor{C0}{\rho_{i\alpha}} - \mathcolor{C1}{\mu_{i\alpha}^2}}}$ elementwise. Note that $\mathcolor{C1}{\mu_{i0}} = f_i$ and $\mathcolor{C2}{\nu_{i0}} = 0$. 
\textbf{\sffamily C.} The local WD values for feature $i$ and scale $\alpha$ are computed elementwise as $\mathcolor{C4}{d_{i\alpha}^{\crowexp}} = \sqrt{\smash[b]{{
    (\mathcolor{C1}{\mu^{\textcolor{black}{\text{\tiny\faCrow}}}_{i\alpha}}-\mathcolor{C1}{\mu^{\textcolor{C3}{\text{\tiny\faCrow}}}_{i\alpha}})^2 
    +
    (\mathcolor{C2}{v^{\textcolor{black}{\text{\tiny\faCrow}}}_{i\alpha}}-\mathcolor{C2}{v^{\textcolor{C3}{\text{\tiny\faCrow}}}_{i\alpha}})^2
    }}}
    $.
They are then spatially averaged, with weights $w_{i\alpha}$ derived from the $\sigma$-map, yielding the scalar WD for feature $i$, $\mathcolor{C4}{d_{i}^{\crowexp}}$. The total WD for the image pair is obtained by adding the contributions from all features.
}
\label{fig:wd_computation_overview}
\end{figure*}

Our experiments are based on the \cc family of methods \citep{ladune2023cool, leguay2023low, leguay2024cool, blard2024overfitted} and more specifically on C3 \citep{C3_2024}, a neural codec which rivals modern classical codecs like VVC~\citep{VVC} in terms of rate--distortion performance, while maintaining a very low computational cost. These codecs are built from three main components: a set of latent arrays, a synthesis network and an entropy network, each of which are optimized per image (see \cref{fig:c3_model_overview}).

The latents consist of a set of multi-resolution arrays, starting with the resolution of the image, followed by progressively smaller resolutions (by a factor of two in each dimension). To synthesize an image, these latents are bilinearly upsampled and concatenated to create a single multi-channel array at the resolution of the image. This array is passed through a synthesis network $f_\theta$, a small convolutional neural network, to output RGB values at every pixel location. The entropy network $g_\psi$ is used to model the conditional distribution of each latent element given its spatial context. The bit sequence is made up of the bits encoding the quantized and entropy coded latents (with probabilities estimated by $g_\psi$) as well as the bits required to store the quantized parameters of both $f_\theta$ and $g_\psi$.

The latent arrays and the parameters of the synthesis and entropy networks are jointly optimized for a rate--distortion objective, with distortion given by the MSE between the original and reconstructed image pixels, and rate approximated by the cross entropy of the latent arrays under the entropy model. Since the latent arrays are quantized, this optimization is done in a quantization-aware manner. For further details on the architecture and optimization, we refer the reader to \citet{ladune2023cool} and \citet{C3_2024}.

In this paper, we use the identical architecture and optimization procedure as C3~\cite{C3_2024}, except for two changes:
1. We supply common randomness arrays to the decoder in the form of pseudo-random i.i.d. standard Gaussian noise, which remains fixed during optimization and decoding. Since the seed is fixed, no additional information needs to be encoded.
2. We replace MSE by the Wasserstein Distortion between the original and its reconstruction.

\section{Wasserstein Distortion}

Wasserstein Distortion (WD)~\citep{qiu-2024-wasserstein-distortion} is a distance metric between pairs of images which, similar to the popular LPIPS~\citep{zhang-2018-the-unreasonable-effectiveness} and DISTS~\citep{ding-2020-image-quality} metrics, can utilize feature embeddings. The feature space is interpreted as a \emph{perceptual} space, in which distances are more predictive of subjective human notions of distance between images than distances between pixel intensities. While WD itself is agnostic to the feature space, we chose to rely on the well-known VGG embeddings~\citep{simonyan2014very}, derived from training a convolutional neural network on the ImageNet task, and also used by LPIPS.

In contrast to LPIPS and DISTS, WD generalizes the notion of pointwise distances in a feature space by taking into account \emph{foveation} and \emph{peripheral vision}. WD is parameterized by a spatial $\sigma$-map of the same resolution as the images. Assume feature maps $f_i$ also have that resolution. Then the local WD for feature $i$ at a location $(x, y)$ is the 2-Wasserstein divergence between the local empirical distributions of $f_i$ for both images, aggregated using a pooling kernel of size $\sigma(x, y)$ and centered at $f_i(x, y)$. For feature maps that have different resolutions than the images, the pooling kernel sizes need to be adjusted accordingly. The WD values are then spatially averaged and summed over all features $i$. As $\sigma$ goes to zero, WD converges to computing pointwise distances, as in LPIPS. This models human vision in the fovea (the center of gaze), which is most accurate in distinguishing deviations between the images. Larger $\sigma$ values model peripheral vision, where humans are capable of distinguishing deviations in texture characteristics, but fail at picking up on subtler deviations. (For a complete description of WD, refer to \citet{qiu-2024-wasserstein-distortion}.)

The larger $\sigma$, the more permissive is WD to \emph{texture resampling}, i.e. replacing a visual texture with one that has similar statistics. For image compression, we exploit this by choosing smaller $\sigma$ in \emph{salient} image regions that are likely to be directly looked at, and larger $\sigma$ in other regions. Making the distance metric more permissive in such regions allows the codec to allocate fewer bits in representing the local image content, while maintaining visual quality. 

Even after approximating empirical distributions as independent Gaussians, computing WD can be computationally costly, since it generally requires aggregation of local statistics with a different pooling kernel at each spatial location. We propose an approximate, but more efficient way of computing WD by discretizing $\sigma$ to powers of two~(\cref{fig:wd_computation_overview}).
For every feature map $f_i$ extracted from VGG (panel~A), we estimate spatial maps of the local first and second raw moments by building a multi-scale cascade akin to a Gaussian pyramid~\cite{burt-1983-the-laplacian-pyramid} using a $3\times3$ convolutional downsampling filter $\mathcal D$ (panel~B). This yields a full spatial map of the local mean $\mu_{i\alpha}$ and standard deviation $\nu_{i\alpha}$ of the $i$th feature, for pooling regions of size $2^\alpha / r_i$, where $\alpha$ is the scale index, and $r_i$ is the relative resolution of the $i$th feature map with respect to the images (e.g., 0.5 if $f_i$ is downsampled by a factor of 2).

\begin{figure*}[t]
\raggedleft
\begin{tikzpicture}[font=\sffamily]
\pgfplotsset{set layers=standard, cell picture=true}
\pgfplotsset{
  tick label style = {font=\sansmath\sffamily},
  every axis label = {font=\sansmath\sffamily},
  legend style = {font=\sansmath\sffamily},
  label style = {font=\sansmath\sffamily}
}
\begin{groupplot}[
  group style={group size=2 by 1, horizontal sep=8pt}, 
  ymin=1200, 
  ymax=2600, 
  ytick={1200,1400,1600,1800,2000,2200,2400,2600},
  scale only axis, 
  height=.22\textheight, 
  width=.45\textwidth,
  enlargelimits={abs=3pt},
  grid,
  axis lines=left,
  legend style={draw,rounded corners,opacity=0.8,font=\footnotesize,cells={anchor=west,opacity=1}},
]
\nextgroupplot[
  xlabel=bits/pixel,
  ylabel=Elo score,
  x tick label style={/pgf/number format/.cd, fixed, precision=2},
  scaled y ticks=base 10:-3,
  xmin=0.0,
  xmax=0.32,
  legend to name=grouplegend,
  reverse legend=true,
  legend columns=2,
  transpose legend=true,
  every axis plot post/.append style={solid, error bars/.cd, y explicit, y dir=both},
  ]
  \draw[-Triangle] (rel axis cs: 0.2, 0.8) -- (rel axis cs: 0.05, 0.95) node[midway, above, sloped] {\footnotesize better};
  \addplot[color=C0, mark=diamond*, mark size=2.5pt]
    table[x=bpp, y=Elo, y error plus expr=\thisrow{p99hi}-\thisrow{Elo}, y error minus expr=\thisrow{Elo}-\thisrow{p99lo}] {
    Elo p99lo p99hi bpp
    1458.8350 1415.2091 1502.4607 0.079
    1848.7307 1816.5380 1880.9236 0.148
    2338.5496 2306.1050 2370.9944 0.387
  }; \addlegendentry{VVC~\citep{VVC}};
  \addplot[color=C4, mark=diamond*, mark size=2.5pt]
    table[x=bpp, y=Elo, y error plus expr=\thisrow{p99hi}-\thisrow{Elo}, y error minus expr=\thisrow{Elo}-\thisrow{p99lo}] {
    Elo p99lo p99hi bpp
    1537.7413 1498.0228 1577.4597 0.078
    1931.5170 1900.5481 1962.4858 0.149
    2264.1577 2233.0432 2295.2722 0.302
  }; \addlegendentry{MLIC+~\citep{jiang2022multi}};
  \addplot[color=C6, mark=diamond*, mark size=2.5pt]
    table[x=bpp, y=Elo, y error plus expr=\thisrow{p99hi}-\thisrow{Elo}, y error minus expr=\thisrow{Elo}-\thisrow{p99lo}] {
    Elo p99lo p99hi bpp
    2138.6208 2094.2341 2183.0076 0.244
    2349.2747 2302.1516 2396.3975 0.424  
  }; \addlegendentry{CDC~\citep{yang-2023-lossy-image}};
  \addplot[color=C2,mark=diamond*, mark size=2.5pt]
    table[x=bpp, y=Elo, y error plus expr=\thisrow{p99hi}-\thisrow{Elo}, y error minus expr=\thisrow{Elo}-\thisrow{p99lo}] {
    Elo p99lo p99hi bpp
    2224.2773 2193.4236 2255.1313 0.142
    2484.2852 2446.6738 2521.8965 0.272
    2698.0610 2642.7197 2753.4023 0.410
  }; \addlegendentry{HiFiC~\citep{mentzer-2020-high-fidelity-generative}};
  \addplot[color=C3, mark=*, mark size=2pt]
    table[x=bpp, y=Elo, y error plus expr=\thisrow{p99hi}-\thisrow{Elo}, y error minus expr=\thisrow{Elo}-\thisrow{p99lo}] {
    Elo p99lo p99hi bpp
    1314.7732 1263.9172 1365.6293 0.078
    1794.4622 1761.7052 1827.2192 0.151
    2186.5173 2156.2515 2216.7832 0.302
  }; \addlegendentry{C3/MSE~\citep{C3_2024}};
  \addplot[color=C1, mark=square*, mark size=2pt]
    table[x=bpp, y=Elo, y error plus expr=\thisrow{p99hi}-\thisrow{Elo}, y error minus expr=\thisrow{Elo}-\thisrow{p99lo}] {
    Elo p99lo p99hi bpp
    1822.3773 1789.8812 1854.8734 0.074
    2187.2837 2156.9400 2217.6272 0.149
    2435.4153 2398.9321 2471.8984 0.294
  }; \addlegendentry{C3/WDs (ours)};
  \node [draw, dashed, shape=rectangle, rounded corners, minimum width=30, minimum height=100] (bpp_selection) at (0.15, 2000) {};
\nextgroupplot[
  yticklabels={\empty},
  xlabel={MACs/pixel at $\approx 0.15$ bits/pixel},
  xmode=log,
  xmin=1e2,
  xmax=1e6,
  every axis plot post/.append style={solid, error bars/.cd, y explicit, y dir=both},
  ]
  \draw[-Triangle] (rel axis cs: 0.2, 0.8) -- (rel axis cs: 0.05, 0.95) node[midway, above, sloped] {\footnotesize better};
  \addplot[color=C0, mark=diamond*, mark size=2.5pt, error bars/.cd, x fixed=500, x dir=both]
    table[x=mpp, y=Elo, y error plus expr=\thisrow{p99hi}-\thisrow{Elo}, y error minus expr=\thisrow{Elo}-\thisrow{p99lo}] {
    Elo p99lo p99hi mpp
    1848.7307 1816.538 1880.9236 1000
  }; \addlegendentry{VVC~\citep{VVC}};
  \addplot[color=C4, mark=diamond*, mark size=2.5pt]
    table[x=mpp, y=Elo, y error plus expr=\thisrow{p99hi}-\thisrow{Elo}, y error minus expr=\thisrow{Elo}-\thisrow{p99lo}] {
    Elo p99lo p99hi mpp
    1931.517 1900.5481 1962.4858 555340
  }; \addlegendentry{MLIC+~\citep{jiang2022multi}};
  \addplot[color=C2,mark=diamond*, mark size=2.5pt]
    table[x=mpp, y=Elo, y error plus expr=\thisrow{p99hi}-\thisrow{Elo}, y error minus expr=\thisrow{Elo}-\thisrow{p99lo}] {
    Elo p99lo p99hi mpp
    2224.2773 2193.4236 2255.1313 609199
  }; \addlegendentry{HiFiC~\citep{mentzer-2020-high-fidelity-generative}};
  \addplot[color=C3, mark=*, mark size=2pt]
    table[x=mpp, y=Elo, y error plus expr=\thisrow{p99hi}-\thisrow{Elo}, y error minus expr=\thisrow{Elo}-\thisrow{p99lo}] {
    Elo p99lo p99hi mpp
    1794.4622 1761.7052 1827.2192 2925
  }; \addlegendentry{C3/MSE~\citep{C3_2024}};
  \addplot[color=C1, mark=square*, mark size=2pt]
    table[x=mpp, y=Elo, y error plus expr=\thisrow{p99hi}-\thisrow{Elo}, y error minus expr=\thisrow{Elo}-\thisrow{p99lo}] {
    Elo p99lo p99hi mpp
    2187.2837 2156.94 2217.6272 3149
  }; \addlegendentry{C3/WDs (ours)};
  \legend{};  
\end{groupplot}
\coordinate (second_plot_border) at (group c2r1.west|-bpp_selection.east);
\draw[dashed, arrows={-Triangle[length=10pt,width=7pt]}] (bpp_selection.east) -- ($(second_plot_border)+(6pt,0)$);
\node[anchor=south] (legend) at (group c2r1.south west){\pgfplotslegendfromname{grouplegend}};
\end{tikzpicture}
\caption{Human rating study results and decoder complexity. \textbf{Left:} Evaluation of image compression methods in terms of visual fidelity  vs. bit rate. Error bars indicate 99th percentile. Squares indicate C3 using CR, circles indicate C3 without CR, and diamonds mark other compression methods. \textbf{Right:} Computational complexity of the decoder of the same methods for the middle bit rate. HiFiC slightly outperforms C3 in terms of visual quality vs. bit rate. However, it requires more than two orders of magnitude more computations at the decoder. Note that neural methods tend to have similar complexity across different bit rates and objectives. This is also true for C3; only the addition of CR increases the complexity slightly. The complexity of VVC can vary across bit rates, and can't strictly be shown in the same plot, since it doesn't use floating point operations; we plot an estimated equivalent on typical hardware~\citep{debargha}.}
\label{fig:elo_and_macs}
\end{figure*}

Local WD maps $d_{i\alpha}$ between the two images are computed elementwise from two sets of $\mu_{i\alpha}$ and $\nu_{i\alpha}$ computed on each image. This gives us dense maps of precomputed local WD distances, but only for specific pooling region sizes corresponding to $\sigma$-values of $2^\alpha / r_i$. We can approximate the WD for a desired $\sigma$-value by interpolating WD values from these precomputed maps (panel~C). First, the $\sigma$-map is adapted for each feature $i$ and scale $\alpha$ by computing
\begin{equation}
\sigma_{i\alpha} = \max\bigl( \mathcal D_{i\alpha} (r_i \sigma), 1\bigr).
\end{equation}
This resizes all pooling regions by a factor $r_i$, proportional to the resolution of $f_i$, and also resamples the map with a resampling operation $D_{i\alpha}$ to match the resolution of $d_{i\alpha}$. We then determine weight maps
\begin{equation}
w_{i\alpha} = \max\bigl(1-\bigl| \log_2 \sigma_{i\alpha} - \alpha \bigr|, 0\bigr)
\end{equation}
which are 1 wherever $\sigma$ is equal to $2^\alpha / r_i$, and linearly fade to 0 as $\sigma$ approaches either $2^{\alpha+1} / r_i$ or $2^{\alpha-1} / r_i$. Each $d_{i\alpha}$ is multiplied with $w_{i\alpha}$, averaged across space, and summed across $\alpha$, to yield one WD value $d_i$ for each feature $i$. Although the precomputed maps for each scale $\alpha$ have differing resolutions, the weights for each spatial location in $f_i$ add up to one, effectively interpolating local WD maps from precomputed WD maps $d_{i\alpha}$. Finally, WD values $d_i$ are aggregated across different features $i$ by summation.

To determine an appropriate $\sigma$-map, we examined two alternatives: First, we may simply chose $\sigma$ constant across the image, which could be interpreted as a belief that every image location is equally likely to be scrutinized by a human observer. In this case the value needs to be chosen conservatively, small enough that a desired level of exact detail in the reconstruction is preserved at all locations. Second, we may derive $\sigma$ from an actual gaze map obtained with an eye tracking device -- or a prediction thereof. In our experiments, we obtained predictions from EML-net~\cite{jia2020eml} in the form of a saliency map $s$ valued between 0 and 1 (\cref{fig:saliency_ablation}). We converted $s$ to a spatial density $p$ using the ad-hoc elementwise formula:
\begin{equation}
p = p_\text{min} + (1 - p_\text{min}) \cdot s \,/\, \overline{s},
\end{equation}
with $p_\text{min} = 0.5$ lower bounding the likelihood, and $\overline{s}$ indicating the spatial mean of $s$. The resulting density $p$ is always positive and averages to one spatially, regardless of the image resolution. We further converted $p$ to $\sigma$ by computing elementwise:
\begin{equation}
\sigma = \sigma_\text{max} \cdot p_\text{min} \,/\, p,
\end{equation}
where $\sigma_\text{max}$ is the maximal value of $\sigma$. This parameter needs to be chosen based on image resolution and viewing distance.

\section{Experimental results}

To evaluate the proposed method, we compared a total of 10 image compression methods on the Challenge on Learned Image Compression (CLIC) 2020 professional validation dataset\footnote{Available at \url{https://data.vision.ee.ethz.ch/cvl/clic/professional_valid_2020.zip}} in a human rater study:
\begin{itemize}
    \item C3, optimized for MSE (C3/MSE)~\citep{C3_2024};
    \item C3, optimized for a weighted version of MSE using $p$ derived from saliency (C3/wMSE);
    \item C3, optimized for MS-SSIM~\citep{wang-2003-multiscale-structural} (C3/MS-SSIM);
    \item C3 with CR, optimized for LPIPS~\citep{zhang-2018-the-unreasonable-effectiveness} (C3/LPIPS);
    \item C3 with CR, optimized for WD using constant $\sigma=8$ (C3/WD8);
    \item C3 with CR, optimized for WD using a $\sigma$-map derived from saliency with $\sigma_\text{max} = 16$ (C3/WDs);
    \item Versatile Video Coding (VVC)~\citep{VVC}, a state-of-the-art commercial image and video codec, as implemented by VTM 23.4 using the intra YUV444 configuration;
    \item MLIC+ \citep{jiang2022multi}, a learned image codec which is state of the art in terms of MSE vs. bit rate;
    \item CDC \citep{yang-2023-lossy-image}, a learned image codec based on diffusion; and
    \item HiFiC \citep{mentzer-2020-high-fidelity-generative}, a learned image codec based on an adversarial loss.
\end{itemize}
For each of the methods, we obtained compressed images and their reconstructions targeting three dataset-average bit rates: 0.075, 0.15, and 0.3 bits/pixel (the same bit rates that CLIC uses). For CDC and HiFiC, we were not able to exactly match these target rates, since it would have required re-training. Still, we included all three available target rates for HiFiC, as well as the two lower target rates for CDC in the study. This gave us a total of 29 method/rate combinations, and hence 29 distinct reconstructions for each of the 41 images in the dataset.

Our evaluation protocol closely follows the CLIC approach. For each individual rating, three images are available to the rater: On one side of the screen, a random $512 \times 432$ pixel crop of the original. On the other, the corresponding crop of two reconstructed images, between which the rater can flip by pressing a key. The rater is asked to select the reconstruction that looks more similar to the original. For 10\% of the ratings, one of the reconstructions is replaced by the original, which is used to assess rater reliability (\emph{golden questions}). While the original is selected at random, the pair of compared methods/rates is dynamically selected to maximize expected information gain considering past ratings.\footnote{We used an open source implementation of the CLIC rating model, available at \url{https://github.com/google-research/google-research/tree/master/elo_rater_model}.} The resulting Elo score for each method/rate minimizes the cross-entropy between observed and predicted ratings.

The main result of our study is shown in \cref{fig:elo_and_macs}. Methods optimized for MSE (C3/MSE, MLIC+, VVC) and the diffusion model CDC all achieve similar trade-offs between bit rate and visual quality, while their complexity varies substantially (with VVC being least complex). HiFiC and C3 optimized for WD achieve much better and comparable trade-offs, but our method requires two orders of magnitude fewer MACs at decoding time.

\begin{figure}[t]
\raggedleft
\begin{tikzpicture}[font=\sffamily]
\pgfplotsset{set layers=standard, cell picture=true}
\pgfplotsset{
  tick label style = {font=\sansmath\sffamily\footnotesize},
  every axis label = {font=\sansmath\sffamily\footnotesize},
  legend style = {font=\sansmath\sffamily\footnotesize},
  label style = {font=\sansmath\sffamily\footnotesize}
}
\begin{axis}[
  group style={group size=2 by 1, horizontal sep=8pt}, 
  ymin=1200, 
  ymax=2600, 
  ytick={1200,1400,1600,1800,2000,2200,2400,2600},
  scale only axis,
  height=.27\textheight, 
  width=.40\textwidth,
  xlabel=bits/pixel,
  ylabel=Elo score,
  x tick label style={/pgf/number format/.cd, fixed, precision=2},
  scaled y ticks=base 10:-3,
  xmin=0.05,
  xmax=0.32,
  reverse legend=true,
  legend pos=south east,
  legend style={draw,rounded corners,opacity=0.8,cells={anchor=west,opacity=1}},
  every axis plot post/.append style={error bars/.cd, y explicit, y dir=both},
  enlargelimits={abs=3pt},
  grid,
  axis lines=left,
]
  \draw[-Triangle] (rel axis cs: 0.2, 0.8) -- (rel axis cs: 0.05, 0.95) node[midway, above, sloped] {\footnotesize better};
  \addplot[color=C3, solid, mark=*, mark size=2pt]
    table[x=bpp, y=Elo, y error plus expr=\thisrow{p99hi}-\thisrow{Elo}, y error minus expr=\thisrow{Elo}-\thisrow{p99lo}] {
    Elo p99lo p99hi bpp
    1314.7732 1263.9172 1365.6293 0.078
    1794.4622 1761.7052 1827.2192 0.151
    2186.5173 2156.2515 2216.7832 0.302
  }; \addlegendentry{C3/MSE};
  \addplot[color=C3!75!black, densely dashed, mark=*, mark size=2pt]
    table[x=bpp, y=Elo, y error plus expr=\thisrow{p99hi}-\thisrow{Elo}, y error minus expr=\thisrow{Elo}-\thisrow{p99lo}] {
    Elo p99lo p99hi bpp
    1421.4607 1376.0242 1466.8972 0.077
    1838.8888 1806.6022 1871.1754 0.151
    2230.5005 2199.7979 2261.2031 0.305  
  }; \addlegendentry{C3/wMSE};
  \addplot[color=C3!50!black, densely dotted, mark=*, mark size=2pt]
    table[x=bpp, y=Elo, y error plus expr=\thisrow{p99hi}-\thisrow{Elo}, y error minus expr=\thisrow{Elo}-\thisrow{p99lo}] {
    Elo p99lo p99hi bpp
    1410.8109 1365.3926 1456.2291 0.075
    1854.9518 1822.7004 1887.2032 0.154
    2162.6143 2131.6013 2193.6270 0.302
  }; \addlegendentry{C3/MS-SSIM};
  \addplot[color=C1!50!black, densely dashed, mark=square*, mark size=2pt]
    table[x=bpp, y=Elo, y error plus expr=\thisrow{p99hi}-\thisrow{Elo}, y error minus expr=\thisrow{Elo}-\thisrow{p99lo}] {
    Elo p99lo p99hi bpp
    1565.6017 1496.4482 1634.7551 0.0796
    1989.3185 1930.8170 2047.8200 0.1551
    2248.7358 2191.6013 2305.8706 0.3020
  }; \addlegendentry{C3/LPIPS};
  \addplot[color=C1!75!black, densely dashdotted, mark=*, mark size=2pt]
    table[x=bpp, y=Elo, y error plus expr=\thisrow{p99hi}-\thisrow{Elo}, y error minus expr=\thisrow{Elo}-\thisrow{p99lo}] {
    Elo p99lo p99hi bpp
    1491.5714 1420.3374 1562.8054 0.0748
    1979.1648 1917.9819 2040.3477 0.1503
    2369.2527 2306.6810 2431.8242 0.3014
  }; \addlegendentry{C3/WD8 (no CR)};
  \addplot[color=C1!75!black, densely dotted, mark=square*, mark size=2pt]
    table[x=bpp, y=Elo, y error plus expr=\thisrow{p99hi}-\thisrow{Elo}, y error minus expr=\thisrow{Elo}-\thisrow{p99lo}] {
    Elo p99lo p99hi bpp
    1797.9539 1764.5658 1831.3419 0.075
    2129.6711 2099.0652 2160.2773 0.150
    2334.6128 2301.3708 2367.8547 0.298
  }; \addlegendentry{C3/WD8};
  \addplot[color=C1, solid, mark=square*, mark size=2pt]
    table[x=bpp, y=Elo, y error plus expr=\thisrow{p99hi}-\thisrow{Elo}, y error minus expr=\thisrow{Elo}-\thisrow{p99lo}] {
    Elo p99lo p99hi bpp
    1822.3773 1789.8812 1854.8734 0.074
    2187.2837 2156.9400 2217.6272 0.149
    2435.4153 2398.9321 2471.8984 0.294
  }; \addlegendentry{C3/WDs};
\end{axis}
\end{tikzpicture}
\caption{Human rating study results for alternative perceptual objectives, including WD without saliency. Error bars indicate 99th percentile. As loss functions, both MS-SSIM and LPIPS lead to instabilities. To achieve reasonable results, we clipped the gradients of the exponentiation operations in MS-SSIM (C3/MS-SSIM), and used a loss equally weighting LPIPS and MSE (C3/LPIPS). We also assessed the effect of weighting MSE using saliency (C3/wMSE).}
\label{fig:elo_ablations}
\end{figure}

\begin{figure}[t]
\pgfplotstableread[col sep = comma]{
method,networks,grid0,grid1,grid2,grid3,grid4,grid5,grid6
MSE,0.015362663875992708,0.009434232617809667,0.0197714767644179,0.015837779333928555,0.007108957728189303,0.003391487927107913,0.0019974318395436902,0.0012447479487646643
WD8 no CR,0.015796443265749187,0.005497763404203331,0.017376513935898136,0.010381237317335767,0.013919215342711384,0.00801314718498871,0.0027158455639259846,0.0011640553215018859
WD8 with CR,0.01655950896987101,0.0010756863994059824,0.021893005219015587,0.01097181785238407,0.012351124064724257,0.008376231479526656,0.0027881032693563255,0.0011332114397479994
}\datalow
\pgfplotstableread[col sep = comma]{ method,networks,grid0,grid1,grid2,grid3,grid4,grid5,grid6
MSE,0.015865253284573555,0.04927170642679168,0.04019958089196646,0.026411424027528704,0.00964165740169403,0.0054419661457536786,0.0027939556351090533,0.00133523384177285
WD8 no CR,0.01611493209876666,0.022843570165938083,0.05064995996881186,0.02919683033009855,0.014960373501952102,0.01161704148824622,0.003632912673510429,0.0013407660682299516
WD8 with CR,0.016616361988026934,0.010484085447171228,0.05955491379675705,0.030631596727932735,0.01569579321345905,0.011904490089452848,0.0037708516096378246,0.001334439241895225
}\datamiddle
\pgfplotstableread[col sep = comma]{
method,networks,grid0,grid1,grid2,grid3,grid4,grid5,grid6
MSE,0.016504406020408702,0.14594381801619372,0.07486647849039334,0.03648492019260075,0.014507240144445039,0.008381837252073171,0.003827246973609052,0.0016344164969490432
WD8 no CR,0.016280596939528824,0.09550030057023211,0.0967683023432406,0.04567279544000218,0.0248301940992838,0.01583847512559193,0.004881202788432923,0.001584410437428188
WD8 with CR,0.01680758649983055,0.06954308359304488,0.11385055031718277,0.04899651616266588,0.025748709602872046,0.016186037373433753,0.005243233581096298,0.0015792368354684695
}\datahigh
\begin{tikzpicture}
\pgfplotsset{
  tick label style = {font=\sansmath\sffamily\footnotesize},
  every axis label = {font=\sansmath\sffamily\footnotesize},
  legend style = {font=\sansmath\sffamily\footnotesize},
  label style = {font=\sansmath\sffamily\footnotesize}
}
\begin{groupplot}[
  group style={group size=3 by 1},
  height=.7\linewidth,
  width=.31\linewidth,
  ybar stacked,
  ymin=0,
  xmin=-0.5,
  xmax=2.5,
  xtick=data,
  xtick style={draw=none},
  ytick align=outside,
  ytick pos=left,
  ymajorgrids=true,
  legend style={cells={anchor=west}, legend pos=outer north east},
  reverse legend=true,
  yticklabel style={
    /pgf/number format/fixed,
    /pgf/number format/fixed zerofill=true,
    /pgf/number format/precision=3,
  },
  scaled y ticks=false,
  x tick label style={rotate=45,anchor=north east},
  legend style={draw=none},
  every axis plot/.style={mark=none,line width=0.5pt},
]
\nextgroupplot[
  xticklabels from table={\datalow}{method},
  bar width=7pt,
  ylabel={bits/pixel},
  ytick={0,0.015,...,0.076},
  yticklabel style={/pgf/number format/precision=3},
  ymax=0.078,
]
\addplot [darkgray, fill=gray] table [y=networks, meta=method, x expr=\coordindex] {\datalow};
\addplot [C0dark, fill=C0light] table [y=grid0, meta=method, x expr=\coordindex] {\datalow};
\addplot [C1dark, fill=C1light] table [y=grid1, meta=method, x expr=\coordindex] {\datalow};
\addplot [C2dark, fill=C2light] table [y=grid2, meta=method, x expr=\coordindex] {\datalow};
\addplot [C3dark, fill=C3light] table [y=grid3, meta=method, x expr=\coordindex] {\datalow};
\addplot [C4dark, fill=C4light] table [y=grid4, meta=method, x expr=\coordindex] {\datalow};
\addplot [C5dark, fill=C5light] table [y=grid5, meta=method, x expr=\coordindex] {\datalow};
\addplot [C6dark, fill=C6light] table [y=grid6, meta=method, x expr=\coordindex] {\datalow};
\nextgroupplot[
  xticklabels from table={\datamiddle}{method},
  bar width=7pt,
  ytick={0,0.03,...,0.151},
  yticklabel style={/pgf/number format/precision=2},
  ymax=0.156,
]
\addplot [darkgray, fill=gray] table [y=networks, meta=method, x expr=\coordindex] {\datamiddle};
\addplot [C0dark, fill=C0light] table [y=grid0, meta=method, x expr=\coordindex] {\datamiddle};
\addplot [C1dark, fill=C1light] table [y=grid1, meta=method, x expr=\coordindex] {\datamiddle};
\addplot [C2dark, fill=C2light] table [y=grid2, meta=method, x expr=\coordindex] {\datamiddle};
\addplot [C3dark, fill=C3light] table [y=grid3, meta=method, x expr=\coordindex] {\datamiddle};
\addplot [C4dark, fill=C4light] table [y=grid4, meta=method, x expr=\coordindex] {\datamiddle};
\addplot [C5dark, fill=C5light] table [y=grid5, meta=method, x expr=\coordindex] {\datamiddle};
\addplot [C6dark, fill=C6light] table [y=grid6, meta=method, x expr=\coordindex] {\datamiddle};
\nextgroupplot[
  xticklabels from table={\datahigh}{method},
  bar width=7pt,
  ytick={0,0.06,...,0.31},
  yticklabel style={/pgf/number format/precision=2},
  ymax=0.312,
]
\addplot [darkgray, fill=gray] table [y=networks, meta=method, x expr=\coordindex] {\datahigh};
\addlegendentry{networks};
\addplot [C0dark, fill=C0light] table [y=grid0, meta=method, x expr=\coordindex] {\datahigh};
\addlegendentry{array 1};
\addplot [C1dark, fill=C1light] table [y=grid1, meta=method, x expr=\coordindex] {\datahigh};
\addlegendentry{array 2};
\addplot [C2dark, fill=C2light] table [y=grid2, meta=method, x expr=\coordindex] {\datahigh};
\addlegendentry{array 3};
\addplot [C3dark, fill=C3light] table [y=grid3, meta=method, x expr=\coordindex] {\datahigh};
\addlegendentry{array 4};
\addplot [C4dark, fill=C4light] table [y=grid4, meta=method, x expr=\coordindex] {\datahigh};
\addlegendentry{array 5};
\addplot [C5dark, fill=C5light] table [y=grid5, meta=method, x expr=\coordindex] {\datahigh};
\addlegendentry{array 6};
\addplot [C6dark, fill=C6light] table [y=grid6, meta=method, x expr=\coordindex] {\datahigh};
\addlegendentry{array 7};
\end{groupplot}
\end{tikzpicture}
\caption{Dataset-average bit allocation into individual latent arrays of varying resolution from array~1 (highest resolution) to array~7 (lowest resolution) for three target bit rates (0.075, 0.15, and 0.3 bits/pixel), and three variants of C3: optimized for MSE~\cite{C3_2024}, and optimized for WD with $\sigma = 8$, supplied with CR or without.}
\label{fig:bit_allocation}
\end{figure}

\begin{figure*}
\centering
\begin{tikzpicture}[
  spy using outlines={rectangle, red, magnification=3, size=2cm, connect spies},
  every node/.style={font=\small},
]
\node[inner sep=0pt] (original)
  {\includegraphics[width=.33\linewidth]{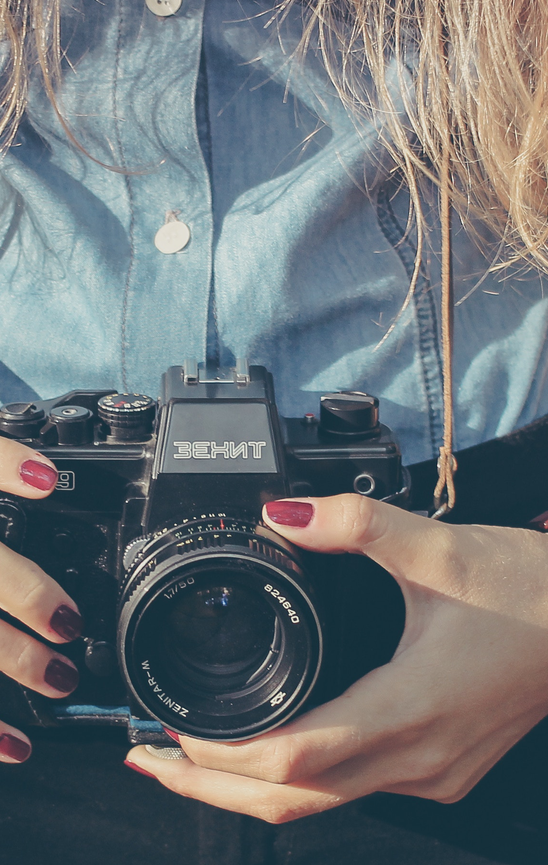}};
\node[inner sep=0pt, right=2pt of original] (mse)
  {\includegraphics[width=.33\linewidth]{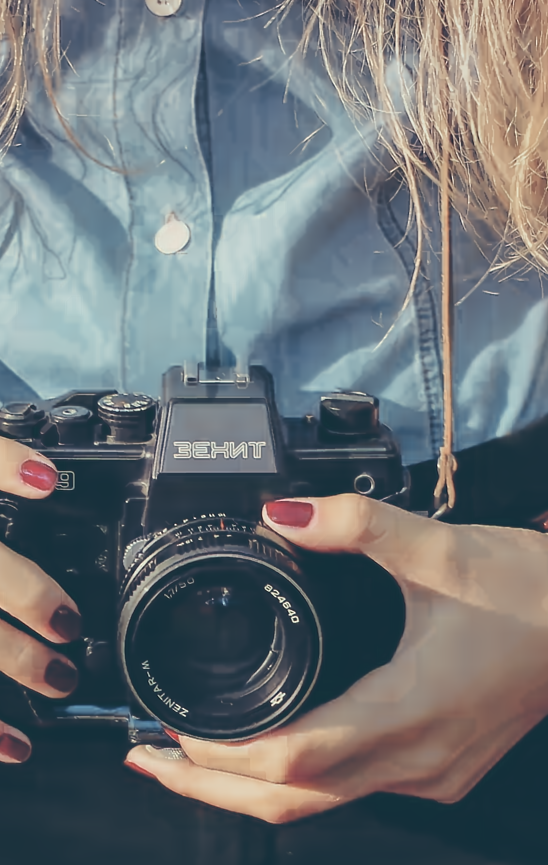}};
\node[inner sep=0pt, right=2pt of mse] (wd_flat)
  {\includegraphics[width=.33\linewidth]{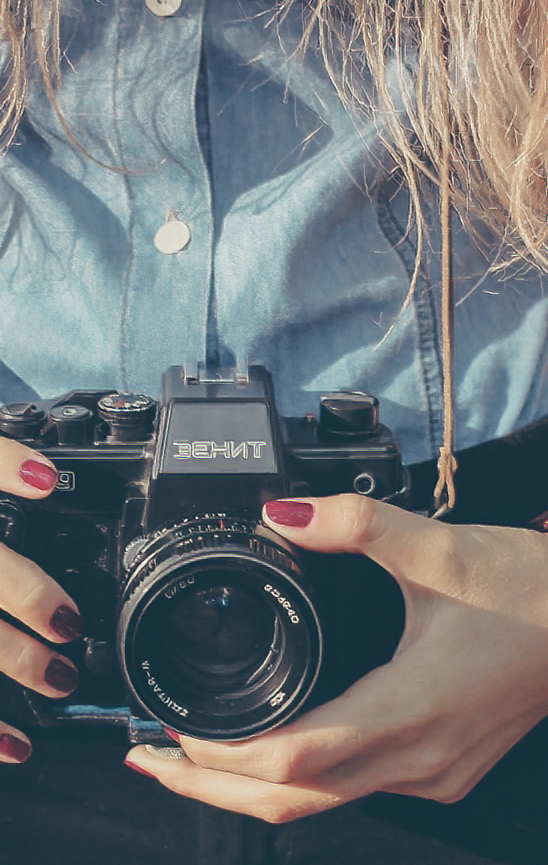}};
\node[inner sep=0pt, below=2pt of original] (saliency)
  {\includegraphics[width=.33\linewidth]{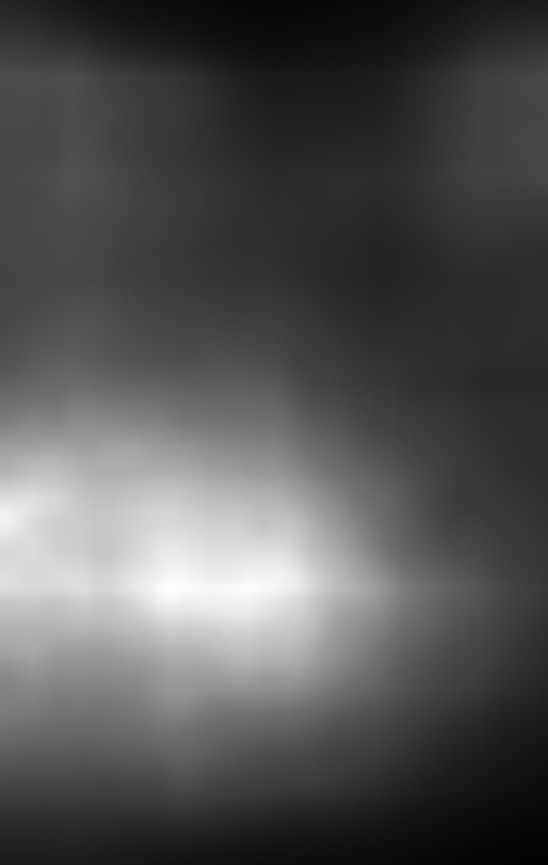}};
\node[inner sep=0pt, below=2pt of mse] (mse_sw)
  {\includegraphics[width=.33\linewidth]{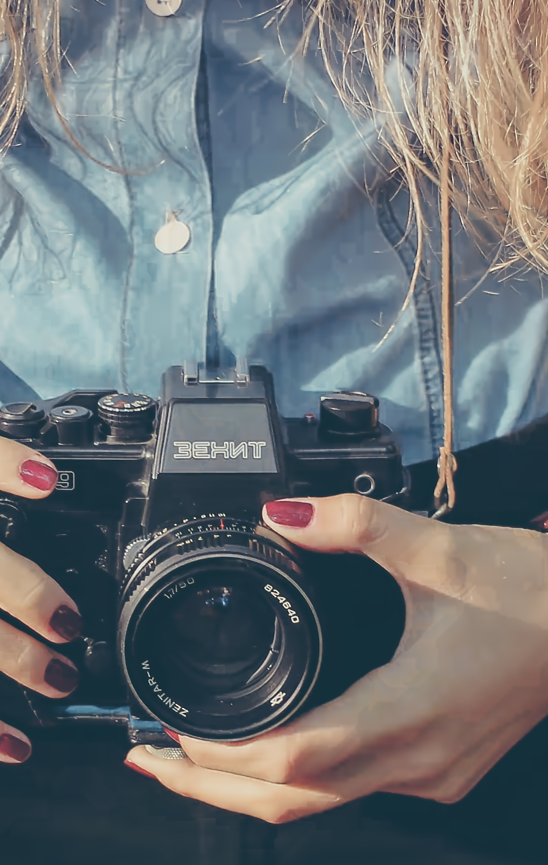}};
\node[inner sep=0pt, below=2pt of wd_flat] (wd_sal)
  {\includegraphics[width=.33\linewidth]{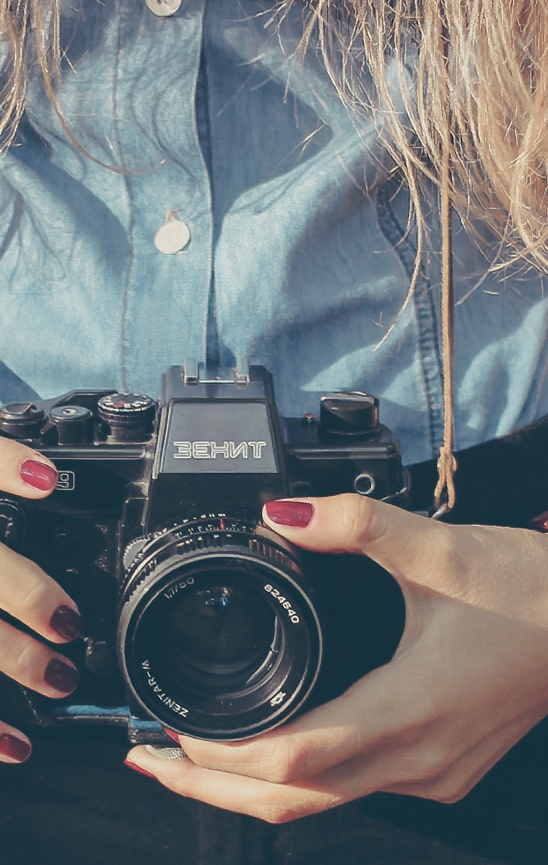}};

\node[above right=0 of original.south west, color=white] {\sffamily original};
\node[above right=0 of mse.south west, color=white] {\sffamily \textcolor{white}{C3/MSE~\citep{C3_2024} at 0.191\,bits/pixel}};
\node[above right=0 of wd_flat.south west, color=white] {\sffamily C3/WD8 at 0.194\,bits/pixel};
\node[above right=0 of saliency.south west, color=white] {\sffamily saliency map};
\node[above right=0 of mse_sw.south west, color=white] {\sffamily C3/wMSE at 0.202\,bits/pixel};
\node[above right=0 of wd_sal.south west, color=white] {\sffamily C3/WDs at 0.191\,bits/pixel};

\node (original_spy1) at ({$(original.west)!.3!(original.east)$} |- {$(original.south)!.205!(original.north)$}) {};
\spy on (original_spy1) in node[below right=1pt] at (original.north west);
\node (mse_spy1) at ({$(mse.west)!.3!(mse.east)$} |- {$(mse.south)!.205!(mse.north)$}) {};
\spy on (mse_spy1) in node[below right=1pt] at (mse.north west);
\node (wd_flat_spy1) at ({$(wd_flat.west)!.3!(wd_flat.east)$} |- {$(wd_flat.south)!.205!(wd_flat.north)$}) {};
\spy on (wd_flat_spy1) in node[below right=1pt] at (wd_flat.north west);
\node (mse_sw_spy1) at ({$(mse_sw.west)!.3!(mse_sw.east)$} |- {$(mse_sw.south)!.205!(mse_sw.north)$}) {};
\spy on (mse_sw_spy1) in node[below right=1pt] at (mse_sw.north west);
\node (wd_sal_spy1) at ({$(wd_sal.west)!.3!(wd_sal.east)$} |- {$(wd_sal.south)!.205!(wd_sal.north)$}) {};
\spy on (wd_sal_spy1) in node[below right=1pt] at (wd_sal.north west);

\node (original_spy2) at ({$(original.west)!.5!(original.east)$} |- {$(original.south)!.12!(original.north)$}) {};
\spy on (original_spy2) in node[below left=1pt] at (original.north east);
\node (mse_spy2) at ({$(mse.west)!.5!(mse.east)$} |- {$(mse.south)!.12!(mse.north)$}) {};
\spy on (mse_spy2) in node[below left=1pt] at (mse.north east);
\node (wd_flat_spy2) at ({$(wd_flat.west)!.5!(wd_flat.east)$} |- {$(wd_flat.south)!.12!(wd_flat.north)$}) {};
\spy on (wd_flat_spy2) in node[below left=1pt] at (wd_flat.north east);
\node (mse_sw_spy2) at ({$(mse_sw.west)!.5!(mse_sw.east)$} |- {$(mse_sw.south)!.12!(mse_sw.north)$}) {};
\spy on (mse_sw_spy2) in node[below left=1pt] at (mse_sw.north east);
\node (wd_sal_spy2) at ({$(wd_sal.west)!.5!(wd_sal.east)$} |- {$(wd_sal.south)!.12!(wd_sal.north)$}) {};
\spy on (wd_sal_spy2) in node[below left=1pt] at (wd_sal.north east);

\end{tikzpicture}
\caption{Crop of image 32 from the CLIC2020 professional dataset (best viewed on screen). \textbf{Top left:} original image. \textbf{Bottom left:} Saliency map $s$ as derived from EML-net (0: black, 1: white). \textbf{Top center:} C3 optimized for MSE, compressed to 0.191 bits/pixel. \textbf{Bottom center:} C3 optimized for MSE, weighted with the density map $p$, compressed to 0.202 bits/pixel. \textbf{Top right:} C3 optimized for WD with $\sigma=8$, compressed to 0.194 bits/pixel. \textbf{Bottom right:} C3 optimized for WD with $\sigma$ derived from the saliency map, compressed to 0.191 bits/pixel. Optimizing for MSE leads to flattened texture on the shirt and skin (center column), while WD retains an accurate perception of texture (right column). However, assuming a flat sigma map in WD is equivalent to assuming all image locations will be perceived via peripheral vision, meaning that the text on the camera is subject to texture resampling, making for instance the text on the lens (``ZENITAR-M'') indecipherable (top right). This is clearly undesirable, as text is a semantically relevant feature. According to the predicted saliency, the camera will be the subject of visual scrutiny. Modulating $\sigma$ according to the saliency map accounts for this. The end result is an image with both improved texture reproduction as well as legible text (bottom right). Note that simply weighting MSE using predicted saliency does not improve texture reproduction (bottom center).}
\label{fig:saliency_ablation}
\end{figure*}

We provide further comparisons in \cref{fig:elo_ablations}, including variants of C3 optimized for other perceptual metrics, and ablations for using saliency and common randomness.\footnote{The CR ablation was added to the study after the others. The corresponding ratings did not contribute to the analysis in \cref{tab:predictivity}.} The results show that optimizing for WD significantly outperforms other metrics in terms of human preference, especially if saliency and CR are used. Removing saliency results in a drop in visual quality at high bit rates, while removing common randomness mostly affects low bit rates. \cref{fig:bit_allocation} reveals that both switching the objective from MSE to WD, and providing CR to a WD-optimized codec lead to a smaller fraction of the bit rate being allocated to the highest-resolution array (array 1), which encodes detail, allowing other arrays to encode more information. This is in line with the intuition that both steps enable texture resampling: switching the objective allows the encoder to find instantiations of texture that are visually equivalent, but take fewer bits to encode with the C3 architecture (for example, the roof in \cref{fig:cr_ablation}), and providing CR enables certain stochastic textures to be reproduced by a form of \enquote{noise shaping} (the grass in \cref{fig:cr_ablation}).

\begin{table*}[t]
\begin{minipage}[t]{0.4\linewidth}
\vspace{0pt}%
\raggedright
\small
\rowcolors{2}{verylightgray}{white}
\begin{tabular}{lCCC}
\toprule
& \text{\% correct} & \text{PCC} & \text{SRCC} \\
\midrule
PSNR & 61.27\% & 0.363 & 0.298 \\
MS-SSIM \cite{wang-2003-multiscale-structural} & 64.64\% & 0.540 & 0.601 \\
NLPD \cite{laparra-2016-perceptual-image} & 64.35\% & 0.539 & 0.606 \\
LPIPS \cite{zhang-2018-the-unreasonable-effectiveness} & 69.96\% & 0.711 & 0.699 \\
DISTS \cite{ding-2020-image-quality} & 67.35\% & 0.726 & 0.724 \\
PIM-5 \cite{bhardwaj-2020-an-unsupervised-information-theoretic} & 69.83\% & 0.764 & 0.785 \\
WD2 ($\sigma=2$) & 72.84\% & - & - \\
WDs ($\sigma_\text{max}=4$) & \textbf{72.87\%} & - & - \\
WD8 ($\sigma=8$) & - & 0.936 & 0.902 \\
WDs ($\sigma_\text{max}=16$) & - & \textbf{0.942} & \textbf{0.913} \\
\bottomrule
\end{tabular}
\end{minipage}\hfill
\begin{minipage}[t]{0.57\linewidth}
\vspace{0pt}%
\caption{Perceptual metric predictivity of human ratings collected throughout our study. \textbf{Left column:} Percentage of 16\,659 binary ratings predicted correctly by each metric. Raters compared randomly selected image crops of $512 \times 432$ pixels. $\sigma$ needs to chosen based on display resolution, and the crops are displayed on screen with $\approx 4$ times lower resolution than the full images (which roughly have QXGA format). Hence, we chose 4 times smaller $\sigma$ values for WD to make crop predictions. \textbf{Middle and right column:} Correlation of metrics with Elo scores. We computed averaged metric values between all full reconstructions and their originals, yielding one metric value for each metric and each of 29 methods/rate combinations. We then assessed how well each metric predicted the 29 scores by computing the Pearson correlation coefficient (PCC) and Spearman's rank correlation coefficient (SRCC) between Elo scores and metric averages.}
\label{tab:predictivity}
\end{minipage}
\end{table*}

\cref{fig:elo_ablations} also reveals that basing WD or MSE on saliency predictions gives both a moderate boost in terms of Elo scores, but the visual improvement with MSE is not sufficient to make it competitive with better metrics. We find that it can substantially improve the quality of semantically important features such as text (\cref{fig:saliency_ablation}). Additional visual examples are provided in the supplement.

After observing the strong performance of WD as an optimization objective, we investigated how well it does as an image quality assessment (IQA) method. \cref{tab:predictivity} shows that WD well outperforms existing metrics both in terms of predicting individual ratings on crops, as well as predicting the Elo rankings of methods/rates on full images. To put the result of nearly 73\% into perspective, note that agreement \emph{among raters} on individual binary answers can vary around 80\%. Remarkably, with a PCC of 94\%, WD is an excellent \emph{linear} predictor of Elo scores, which we did not anticipate.

\section{Discussion}

Image and video compression applications enforce draconian limitations on computational complexity, so much so that codecs using generative methods are generally infeasible to deploy on today's (and even tomorrow's) mobile devices, for reasons of both latency and energy efficiency. Neural overfitted codecs were introduced in COIN \citep{dupont2021coin} and NeRV \citep{chen2021nerv} for images and video, respectively, introducing a new class of neural image compression methods that mirror the asymmetric computational complexity trade-off between the encoder (high complexity) and decoder (low complexity) of hybrid video codecs such as VVC~\citep{VVC}. Initial follow up work focused on reducing the encoding time \citep{strumpler2022implicit, schwarz2022meta} or extending the capabilities of these codecs to various data modalities \citep{dupont2022coin++, takikawa2022variable, huang2022compressing}. Better architectures and more refined quantization and domain specific optimizations have improved compression performance both for images \citep{schwarz2023modality, guo2023compression, he2023recombiner, gordon2023quantizing} and video \citep{lee2023ffnerv, maiya2023nirvana, kwan2023hinerv, kwan2024nvrc, gao2024pnvc}. However, many of these improvements come at the cost of increased computational complexity for decoding, weakening one of the key advantages of overfitted codecs in the first place. \cc \citep{ladune2023cool} and follow-up work \citep{leguay2023low, leguay2024cool, blard2024overfitted} was introduced later, maintaining the low decoding cost of COIN and improving performance by not only learning a decoder network, but also latent arrays and an entropy model per image. C3 \citep{C3_2024} further improved performance in terms of MSE vs. bit rate.

In the present work, we introduce a neural image compression method which is nearly the same as C3, but optimized for Wasserstein Distortion. We evaluate it using a human rater study, which reveals that the method truly is good, cheap, \emph{and} fast! To our knowledge, this is the first method to achieve a comparable quality--rate trade-off at this decoding complexity. It provides a compelling demonstration that generative methods are not necessary to achieve highly efficient image compression. With that said, encoding complexity remains a significant limitation of overfitted compression -- even more so due to the complexity of computing WD instead of MSE during optimization. A naive implementation computing WD for arbitrary pooling sizes $\sigma$ would be computationally prohibitive. Our implementation, while not specifically optimized for speed, still leads to a $\approx$6-fold increase in wall time required for each optimization step.

With regards to the strong performance of WD both as a loss function and IQA method in our study, we can ask why it performs better than LPIPS, even though we also use a feature space derived from VGG. There are three main differences between WD and LPIPS: 1. Our implementation of WD uses raw VGG features rather than fitting linear \enquote{fine-tuning} layers to human ratings. Instead, we include the same VGG features computed on two downsampled versions of the image. This adds the same visual features at larger scales, making the overall set of features more robust to scale effects. 2. To achieve stable results, we had to add MSE to the LPIPS loss at equal weights, which was not necessary with WD. The reason may be that we included the image pixels as a \enquote{0th} layer. Note that this is not equivalent, since 3. WD always compares local statistics (i.e. mean and standard deviation) aggregated according to the $\sigma$-map rather than pixel or feature values directly, which is more in line with models of peripheral vision.

It's worth noting that our choice of feature space and $\sigma$-map is entirely ad-hoc. More work is needed to identify which types of visual features are most useful, and which may be redundant (in the spirit of \citet{portilla-2000-a-parametric-texture}). Better saliency models, perhaps targeted specifically at image compression, could be identified. Lastly, the performance of WD for any and all of these choices should be validated with more extensive human rating studies, for which we can only provide a starting point here.

\section*{Acknowledgements}
We would like to thank Wei Jiang and Ruihan Yang for providing reconstructions at several target bit rates for the MLIC+ and CDC baselines, respectively.

{
    \small
    \bibliographystyle{ieeenat_fullname}
    \bibliography{main}

\begin{thebibliography}{42}
\providecommand{\natexlab}[1]{#1}
\providecommand{\url}[1]{\texttt{#1}}
\expandafter\ifx\csname urlstyle\endcsname\relax
  \providecommand{\doi}[1]{doi: #1}\else
  \providecommand{\doi}{doi: \begingroup \urlstyle{rm}\Url}\fi

\bibitem[VVC(2020)]{VVC}
{ITU-T} rec. {H.266} \& {ISO/IEC} 23090-3: Versatile video coding, 2020.

\bibitem[Ahmed et~al.(1974)Ahmed, Natarajan, and Rao]{ahmed-1974-discrete-cosine}
N. Ahmed, T. Natarajan, and K.~R. Rao.
\newblock Discrete cosine transform.
\newblock \emph{{IEEE} Trans. on Computers}, C-23\penalty0 (1), 1974.

\bibitem[Balas et~al.(2009)Balas, Nakano, and Rosenholtz]{balas-2009-a-summary-statistic-representation}
B. Balas, L. Nakano, and R. Rosenholtz.
\newblock A summary-statistic representation in peripheral vision explains visual crowding.
\newblock \emph{Journal of Vision}, 9\penalty0 (12), 2009.

\bibitem[Bhardwaj et~al.(2020)Bhardwaj, Fischer, Ballé, and Chinen]{bhardwaj-2020-an-unsupervised-information-theoretic}
S. Bhardwaj, I. Fischer, J. Ballé, and T. Chinen.
\newblock An unsupervised information-theoretic perceptual quality metric.
\newblock In \emph{Adv. in Neural Information Processing Systems}, 2020.

\bibitem[Blard et~al.(2024)Blard, Ladune, Philippe, Clare, Jiang, and D{\'e}forges]{blard2024overfitted}
T. Blard, T. Ladune, P. Philippe, G. Clare, X. Jiang, and O. D{\'e}forges.
\newblock Overfitted image coding at reduced complexity.
\newblock \emph{arXiv preprint arXiv:2403.11651}, 2024.

\bibitem[Blau and Michaeli(2018)]{blau-2018-the-perception--distortion-tradeoff}
Y. Blau and T. Michaeli.
\newblock The perception--distortion tradeoff.
\newblock In \emph{2018 {IEEE/CVF} Conf. on Computer Vision and Pattern Recognition}, 2018.

\bibitem[Burt and Adelson(1983)]{burt-1983-the-laplacian-pyramid}
P.~J. Burt and E.~H. Adelson.
\newblock The {Laplacian} pyramid as a compact image code.
\newblock \emph{{IEEE} Trans. on Communications}, 31\penalty0 (4), 1983.

\bibitem[Chen et~al.(2021)Chen, He, Wang, Ren, Lim, and Shrivastava]{chen2021nerv}
H. Chen, B. He, H. Wang, Y. Ren, S.~N. Lim, and A. Shrivastava.
\newblock {NeRV}: Neural representations for videos.
\newblock \emph{Adv. in Neural Information Processing Systems}, 34:\penalty0 21557--21568, 2021.

\bibitem[Dhariwal and Nichol(2021)]{diffusion}
P. Dhariwal and A. Nichol.
\newblock Diffusion models beat gans on image synthesis.
\newblock In \emph{Adv. in Neural Information Processing Systems}, pages 8780--8794, 2021.

\bibitem[Ding et~al.(2022)Ding, Ma, Wang, and Simoncelli]{ding-2020-image-quality}
K. Ding, K. Ma, S. Wang, and E.~P. Simoncelli.
\newblock Image quality assessment: Unifying structure and texture similarity.
\newblock \emph{{IEEE} Trans. on Pattern Analysis and Machine Intelligence}, 44\penalty0 (5), 2022.

\bibitem[Dupont et~al.(2021)Dupont, Goli{\'n}ski, Alizadeh, Teh, and Doucet]{dupont2021coin}
E. Dupont, A. Goli{\'n}ski, M. Alizadeh, Y.~W. Teh, and A. Doucet.
\newblock {COIN}: Compression with implicit neural representations.
\newblock \emph{arXiv preprint arXiv:2103.03123}, 2021.

\bibitem[Dupont et~al.(2022)Dupont, Loya, Alizadeh, Golinski, Teh, and Doucet]{dupont2022coin++}
E. Dupont, H. Loya, M. Alizadeh, A. Golinski, Y.~W. Teh, and A. Doucet.
\newblock {COIN++}: Neural compression across modalities.
\newblock \emph{Trans. on Machine Learning Research}, 2022.

\bibitem[Freeman and Simoncelli(2011)]{freeman-2011-metamers-of-the-ventral}
J. Freeman and E.~P. Simoncelli.
\newblock Metamers of the ventral stream.
\newblock \emph{Nature Neuroscience}, 14\penalty0 (9), 2011.

\bibitem[Gao et~al.(2024)Gao, Kwan, Zhang, and Bull]{gao2024pnvc}
G. Gao, H.~M. Kwan, F. Zhang, and D. Bull.
\newblock {PNVC}: Towards practical inr-based video compression.
\newblock \emph{arXiv preprint arXiv:2409.00953}, 2024.

\bibitem[Goodfellow et~al.(2014)Goodfellow, Pouget-Abadie, Mirza, Xu, Warde-Farley, Ozair, Courville, and Bengio]{gans}
I. Goodfellow, J. Pouget-Abadie, M. Mirza, B. Xu, D. Warde-Farley, S. Ozair, A. Courville, and Y. Bengio.
\newblock Generative adversarial nets.
\newblock In \emph{Adv. in Neural Information Processing Systems}, 2014.

\bibitem[Gordon et~al.(2023)Gordon, Chng, MacDonald, and Lucey]{gordon2023quantizing}
C. Gordon, S.-F. Chng, L. MacDonald, and S. Lucey.
\newblock On quantizing implicit neural representations.
\newblock In \emph{Proc. of the IEEE/CVF Winter Conf. on Applications of Computer Vision}, pages 341--350, 2023.

\bibitem[Guo et~al.(2023)Guo, Flamich, He, Chen, and Hern{\'a}ndez-Lobato]{guo2023compression}
Z. Guo, G. Flamich, J. He, Z. Chen, and J.~M. Hern{\'a}ndez-Lobato.
\newblock Compression with {B}ayesian implicit neural representations.
\newblock \emph{arXiv preprint arXiv:2305.19185}, 2023.

\bibitem[He et~al.(2023)He, Flamich, Guo, and Hern{\'a}ndez-Lobato]{he2023recombiner}
J. He, G. Flamich, Z. Guo, and J.~M. Hern{\'a}ndez-Lobato.
\newblock Recombiner: Robust and enhanced compression with {B}ayesian implicit neural representations.
\newblock \emph{arXiv preprint arXiv:2309.17182}, 2023.

\bibitem[Huang and Hoefler(2023)]{huang2022compressing}
L. Huang and T. Hoefler.
\newblock Compressing multidimensional weather and climate data into neural networks.
\newblock In \emph{The Eleventh Int. Conf. on Learning Representations}, 2023.

\bibitem[Jia and Bruce(2020)]{jia2020eml}
S. Jia and N.~D. Bruce.
\newblock {EML-net}: An expandable multi-layer network for saliency prediction.
\newblock \emph{Image and vision computing}, 95:\penalty0 103887, 2020.

\bibitem[Jiang et~al.(2023)Jiang, Yang, Zhai, Ning, Gao, and Wang]{jiang2022multi}
W. Jiang, J. Yang, Y. Zhai, P. Ning, F. Gao, and R. Wang.
\newblock {MLIC}: Multi-reference entropy model for learned image compression.
\newblock In \emph{Proc. of the 31st ACM Int. Conf. on Multimedia}, page 7618–7627, 2023.

\bibitem[Kim et~al.(2024)Kim, Bauer, Theis, Schwarz, and Dupont]{C3_2024}
H. Kim, M. Bauer, L. Theis, J.~R. Schwarz, and E. Dupont.
\newblock C3: High-performance and low-complexity neural compression from a single image or video.
\newblock In \emph{Proceedings of the IEEE/CVF Conf. on Computer Vision and Pattern Recognition (CVPR)}, pages 9347--9358, 2024.

\bibitem[Kwan et~al.(2023)Kwan, Gao, Zhang, Gower, and Bull]{kwan2023hinerv}
H.~M. Kwan, G. Gao, F. Zhang, A. Gower, and D. Bull.
\newblock {HiNeRV}: Video compression with hierarchical encoding based neural representation, 2023.

\bibitem[Kwan et~al.(2024)Kwan, Gao, Zhang, Gower, and Bull]{kwan2024nvrc}
H.~M. Kwan, G. Gao, F. Zhang, A. Gower, and D. Bull.
\newblock {NVRC}: Neural video representation compression.
\newblock \emph{arXiv preprint arXiv:2409.07414}, 2024.

\bibitem[Ladune et~al.(2023)Ladune, Philippe, Henry, Clare, and Leguay]{ladune2023cool}
T. Ladune, P. Philippe, F. Henry, G. Clare, and T. Leguay.
\newblock {COOL-CHIC}: Coordinate-based low complexity hierarchical image codec.
\newblock In \emph{Proc. of the IEEE/CVF Int. Conf. on Computer Vision}, pages 13515--13522, 2023.

\bibitem[Laparra et~al.(2016)Laparra, Ballé, Berardino, and Simoncelli]{laparra-2016-perceptual-image}
V. Laparra, J. Ballé, A. Berardino, and E.~P. Simoncelli.
\newblock Perceptual image quality assessment using a normalized {Laplacian} pyramid.
\newblock In \emph{Human Vision and Electronic Imaging 2016}, 2016.

\bibitem[Lee et~al.(2023)Lee, Rho, Ko, and Park]{lee2023ffnerv}
J.~C. Lee, D. Rho, J.~H. Ko, and E. Park.
\newblock {FFNeRV}: Flow-guided frame-wise neural representations for videos.
\newblock 2023.

\bibitem[Leguay et~al.(2023)Leguay, Ladune, Philippe, Clare, and Henry]{leguay2023low}
T. Leguay, T. Ladune, P. Philippe, G. Clare, and F. Henry.
\newblock Low-complexity overfitted neural image codec.
\newblock \emph{arXiv preprint arXiv:2307.12706}, 2023.

\bibitem[Leguay et~al.(2024)Leguay, Ladune, Philippe, and D{\'e}forges]{leguay2024cool}
T. Leguay, T. Ladune, P. Philippe, and O. D{\'e}forges.
\newblock {COOL-CHIC} {V}ideo: Learned video coding with 800 parameters.
\newblock In \emph{2024 Data Compression Conf. (DCC)}, pages 23--32. IEEE, 2024.

\bibitem[Maiya et~al.(2023)Maiya, Girish, Ehrlich, Wang, Lee, Poirson, Wu, Wang, and Shrivastava]{maiya2023nirvana}
S.~R. Maiya, S. Girish, M. Ehrlich, H. Wang, K.~S. Lee, P. Poirson, P. Wu, C. Wang, and A. Shrivastava.
\newblock {NIRVANA}: Neural implicit representations of videos with adaptive networks and autoregressive patch-wise modeling.
\newblock In \emph{Proc. of the IEEE/CVF Conf. on Computer Vision and Pattern Recognition}, pages 14378--14387, 2023.

\bibitem[Mentzer et~al.(2020)Mentzer, Toderici, Tschannen, and Agustsson]{mentzer-2020-high-fidelity-generative}
F. Mentzer, G. Toderici, M. Tschannen, and E. Agustsson.
\newblock High-fidelity generative image compression.
\newblock In \emph{Adv. in Neural Information Processing Systems 33}, 2020.

\bibitem[Mukherjee(2024)]{debargha}
D. Mukherjee.
\newblock Private communication, 2024.

\bibitem[Portilla and Simoncelli(2000)]{portilla-2000-a-parametric-texture}
J. Portilla and E.~P. Simoncelli.
\newblock A parametric texture model based on joint statistics of complex wavelet coefficients.
\newblock \emph{Int. Journal of Computer Vision}, 40, 2000.

\bibitem[Qiu et~al.(2024)Qiu, Wagner, Ballé, and Theis]{qiu-2024-wasserstein-distortion}
Y. Qiu, A.~B. Wagner, J. Ballé, and L. Theis.
\newblock Wasserstein distortion: Unifying fidelity and realism.
\newblock In \emph{2024 58th Ann. Conf. on Information Sciences and Systems ({CISS})}, 2024.

\bibitem[Schwarz and Teh(2022)]{schwarz2022meta}
J. Schwarz and Y.~W. Teh.
\newblock Meta-learning sparse compression networks.
\newblock \emph{Trans. on Machine Learning Research}, 2022.

\bibitem[Schwarz et~al.(2023)Schwarz, Tack, Teh, Lee, and Shin]{schwarz2023modality}
J.~R. Schwarz, J. Tack, Y.~W. Teh, J. Lee, and J. Shin.
\newblock Modality-agnostic variational compression of implicit neural representations.
\newblock In \emph{Proc. of the 40th Int. Conf. on Machine Learning}, 2023.

\bibitem[Simonyan and Zisserman(2014)]{simonyan2014very}
K. Simonyan and A. Zisserman.
\newblock Very deep convolutional networks for large-scale image recognition.
\newblock \emph{arXiv preprint arXiv:1409.1556}, 2014.

\bibitem[Str{\"u}mpler et~al.(2022)Str{\"u}mpler, Postels, Yang, Gool, and Tombari]{strumpler2022implicit}
Y. Str{\"u}mpler, J. Postels, R. Yang, L.~V. Gool, and F. Tombari.
\newblock Implicit neural representations for image compression.
\newblock In \emph{European Conf. on Computer Vision}, pages 74--91. Springer, 2022.

\bibitem[Takikawa et~al.(2022)Takikawa, Evans, Tremblay, M{\"u}ller, McGuire, Jacobson, and Fidler]{takikawa2022variable}
T. Takikawa, A. Evans, J. Tremblay, T. M{\"u}ller, M. McGuire, A. Jacobson, and S. Fidler.
\newblock Variable bitrate neural fields.
\newblock In \emph{ACM SIGGRAPH 2022 Conf. Proc.}, pages 1--9, 2022.

\bibitem[Wang et~al.(2003)Wang, Simoncelli, and Bovik]{wang-2003-multiscale-structural}
Z. Wang, E.~P. Simoncelli, and A.~C. Bovik.
\newblock Multiscale structural similarity for image quality assessment.
\newblock In \emph{The Thirty-Seventh Asilomar Conf. on Signals, Systems \& Computers, 2003}, 2003.

\bibitem[Yang and Mandt(2023)]{yang-2023-lossy-image}
R. Yang and S. Mandt.
\newblock Lossy image compression with conditional diffusion models.
\newblock In \emph{Adv. in Neural Information Processing Systems}, 2023.

\bibitem[Zhang et~al.(2018)Zhang, Isola, Efros, Shechtman, and Wang]{zhang-2018-the-unreasonable-effectiveness}
R. Zhang, P. Isola, A.~A. Efros, E. Shechtman, and O. Wang.
\newblock The unreasonable effectiveness of deep features as a perceptual metric.
\newblock In \emph{2018 {IEEE/CVF} Conf. on Computer Vision and Pattern Recognition}, 2018.

\end{thebibliography}
}

\clearpage
\onecolumn
\setcounter{page}{1}

{\centering
\Large
\textbf{\thetitle}\\
\vspace{0.5em}Supplementary Material \\
\vspace{1.0em}
}

\section{All reconstructions and study data}

The original images and all reconstructions in PNG format are available for download at the following locations.

\begin{table}[h!]
\begin{tabular}{lll}
\toprule
method & file size & URL \\
\midrule
original images & 129 MB & \url{https://storage.googleapis.com/wasserstein_c3/original.zip} \\
C3/MSE~\citep{C3_2024} & 216 MB & \url{https://storage.googleapis.com/wasserstein_c3/c3-mse.zip} \\
C3/wMSE & 209 MB & \url{https://storage.googleapis.com/wasserstein_c3/c3-wmse.zip} \\
C3/MS-SSIM & 236 MB & \url{https://storage.googleapis.com/wasserstein_c3/c3-ms-ssim.zip} \\
C3/LPIPS & 403 MB & \url{https://storage.googleapis.com/wasserstein_c3/c3-lpips.zip} \\
C3/WD8 & 439 MB & \url{https://storage.googleapis.com/wasserstein_c3/c3-wd8.zip} \\
C3/WDs & 437 MB & \url{https://storage.googleapis.com/wasserstein_c3/c3-wds.zip} \\
VVC~\citep{VVC} & 379 MB & \url{https://storage.googleapis.com/wasserstein_c3/vvc.zip} \\
MLIC+~\citep{jiang2022multi} & 231 MB & \url{https://storage.googleapis.com/wasserstein_c3/mlicplus.zip} \\
CDC~\citep{yang-2023-lossy-image} & 403 MB & \url{https://storage.googleapis.com/wasserstein_c3/cdc.zip} \\
HiFiC~\citep{mentzer-2020-high-fidelity-generative} & 383 MB & \url{https://storage.googleapis.com/wasserstein_c3/hific.zip} \\
\bottomrule
\end{tabular}
\end{table}

The evaluation data (rater responses, metric evaluations) is available at \url{https://storage.googleapis.com/wasserstein_c3/eval.zip}.

\section{Additional examples}

Please see the next pages for further selected examples.

\begin{landscape}
\begin{figure*}
\includegraphics[width=.5\linewidth]{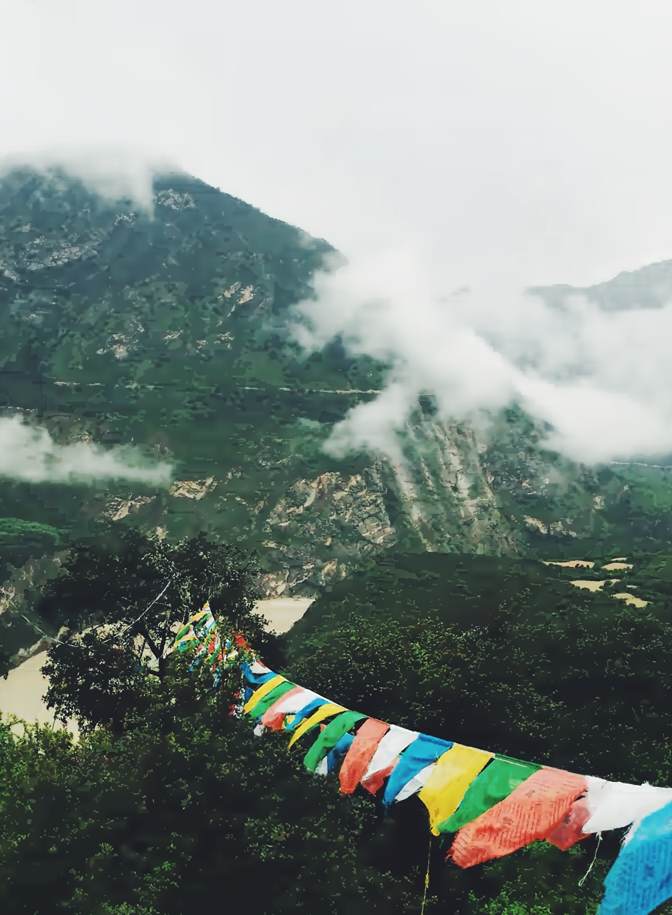}%
\includegraphics[width=.5\linewidth]{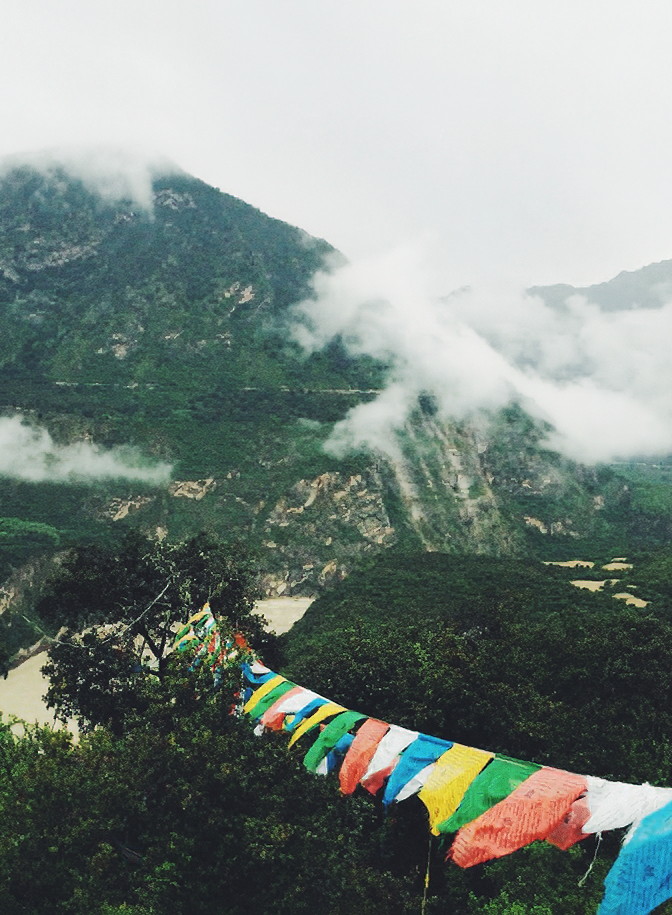}
\caption{Horizontal crop of a image 6 from the CLIC2020 professional dataset (best viewed on screen).
\textbf{Left:} C3 optimized for MSE, compressed to 0.321 bits/pixel.
\textbf{Right:} C3 optimized for WD with $\sigma=8$, compressed to 0.270 bits/pixel.
While optimization for MSE leads to flattened texture and staircasing artifacts, as seen in the reproduction of the vegetation, texture is vastly improved in the WD-optimized version, while using 15\% fewer bits.}
\end{figure*}

\begin{figure*}
\includegraphics[width=.25\linewidth]{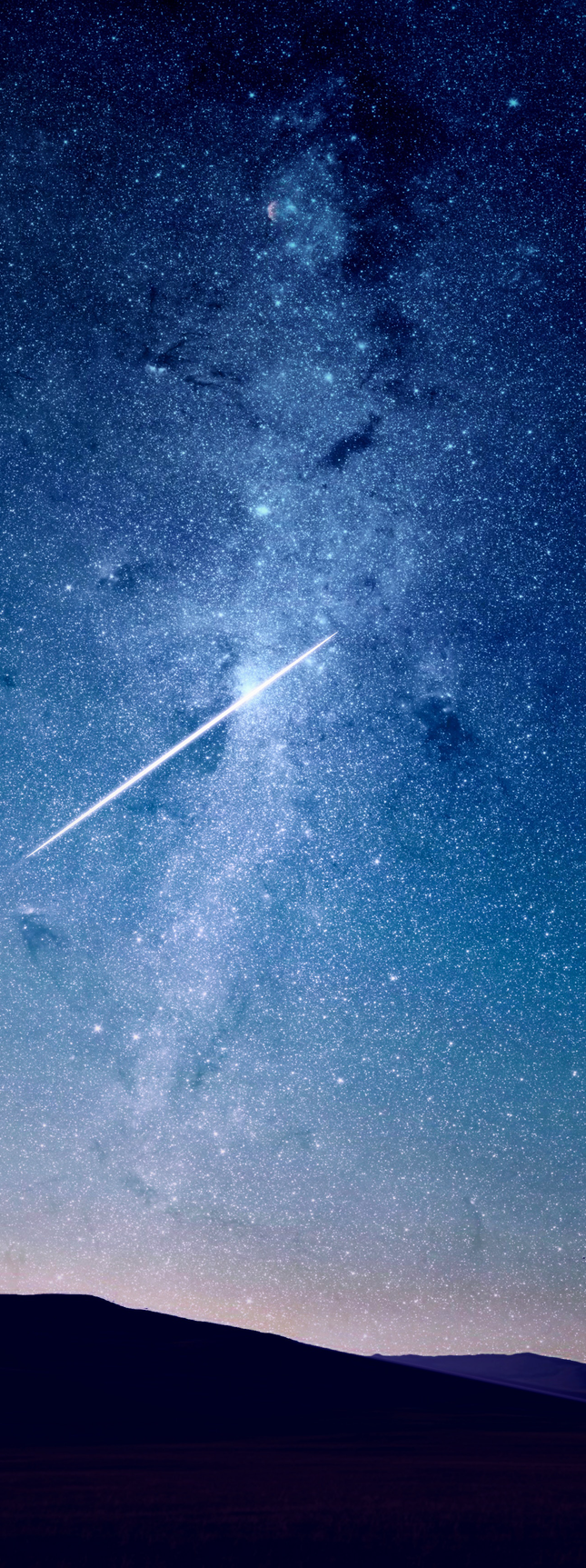}%
\includegraphics[width=.25\linewidth]{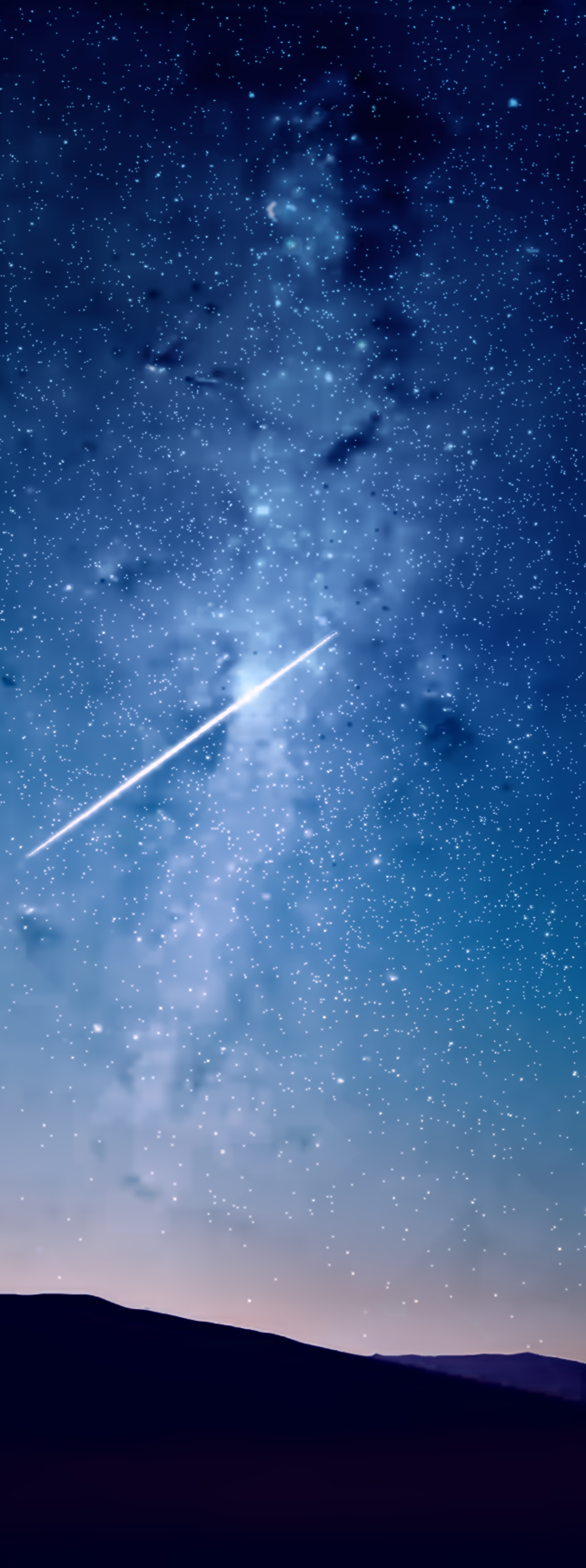}%
\includegraphics[width=.25\linewidth]{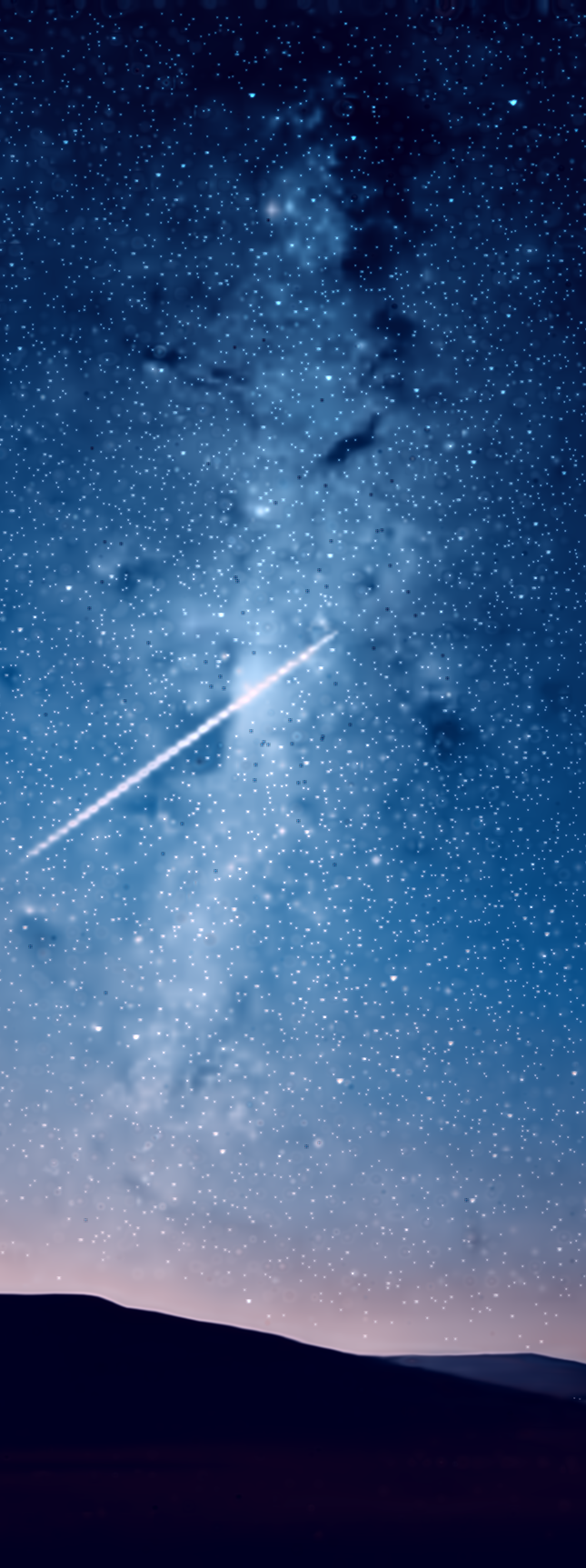}%
\includegraphics[width=.25\linewidth]{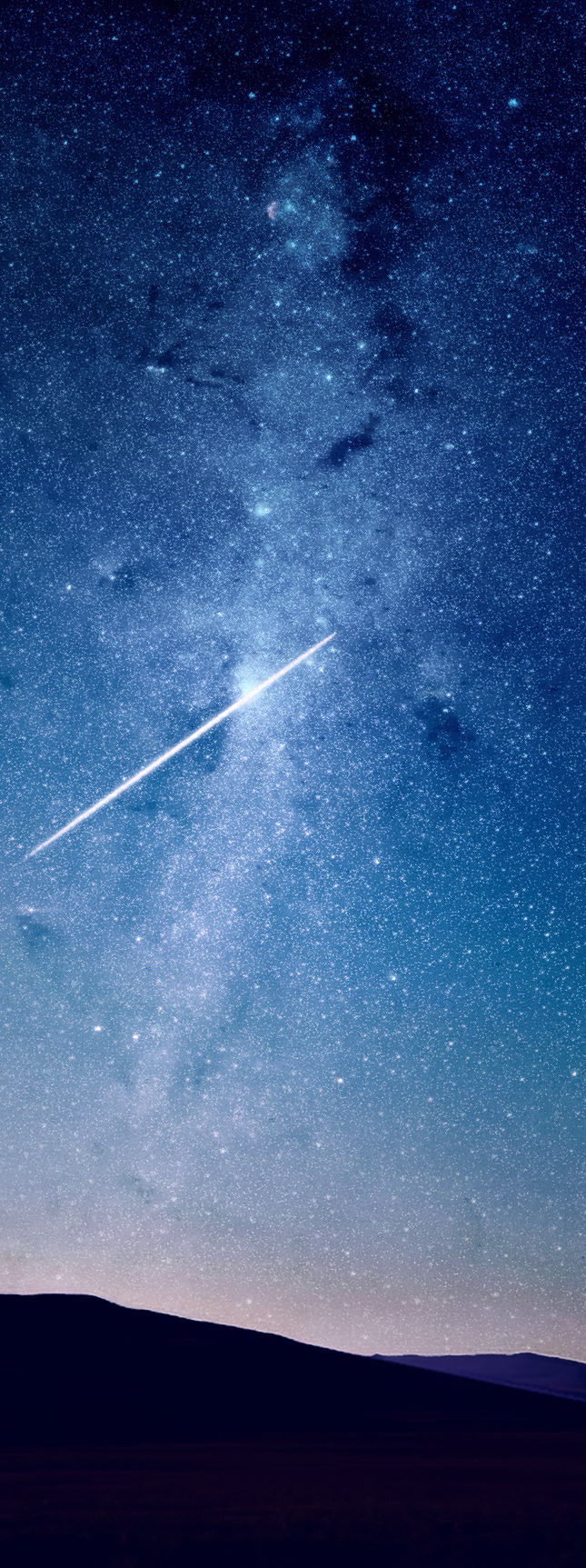}
\caption{Horizontal crop of image 19 from the CLIC2020 professional dataset (best viewed on screen).
\textbf{Far left:} Original image.
\textbf{Center left:} C3 optimized for MSE, compressed to 0.076 bits/pixel.
\textbf{Center right:} C3 optimized for MS-SSIM, compressed to 0.088 bits/pixel.
\textbf{Far right:} C3 optimized for WD with $\sigma=8$, compressed to 0.076 bits/pixel.
Under traditional perceptual metrics such as MSE and MS-SSIM, the visual quality of textures such as the starry night sky suffers, even though MS-SSIM is designed to account for visual contrast masking. Neither of them benefit from common randomness provided by the method. WD accounts much better for perception of visual texture and makes use of CR to achieve a more accurate visual impression.
}
\end{figure*}

\begin{figure*}
\includegraphics[width=.25\linewidth]{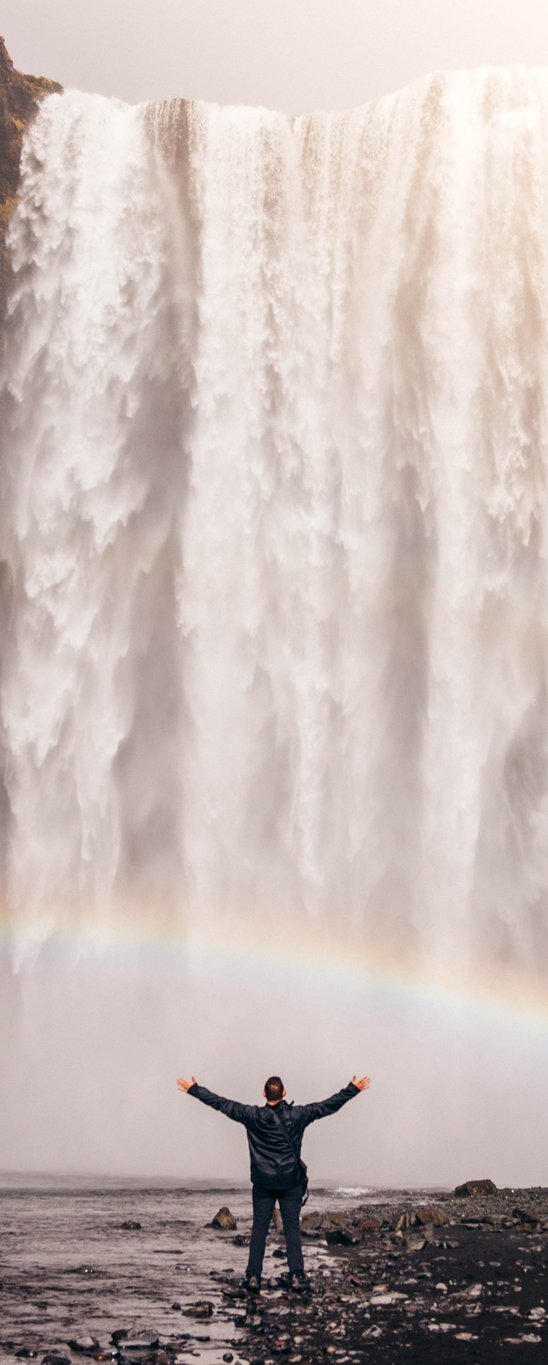}%
\includegraphics[width=.25\linewidth]{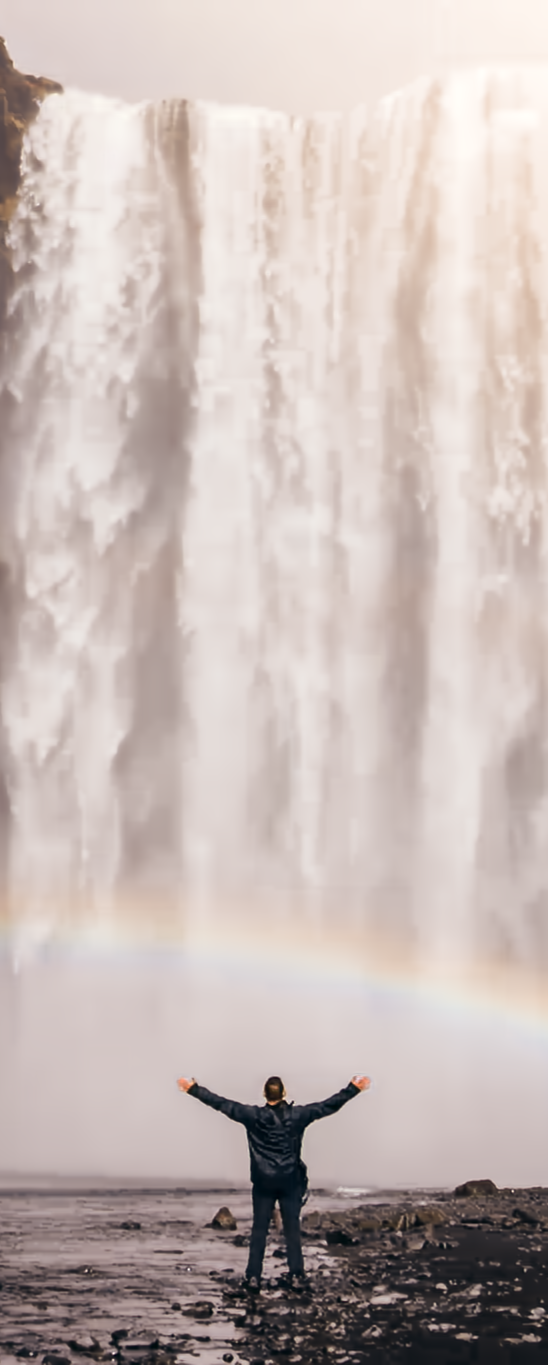}%
\includegraphics[width=.25\linewidth]{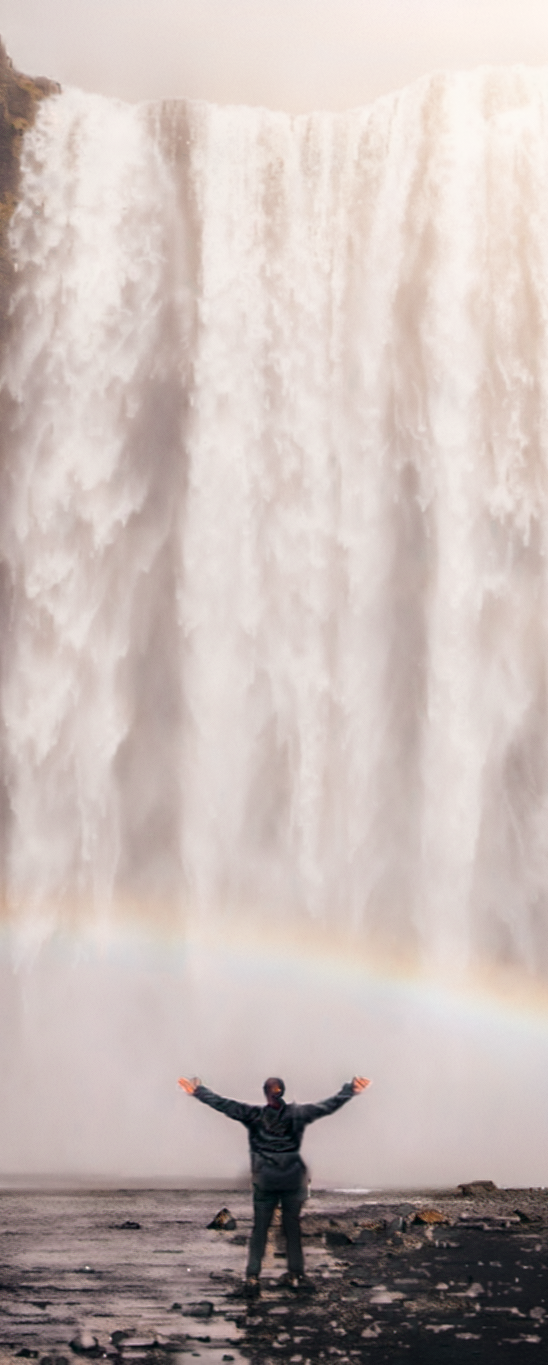}%
\includegraphics[width=.25\linewidth]{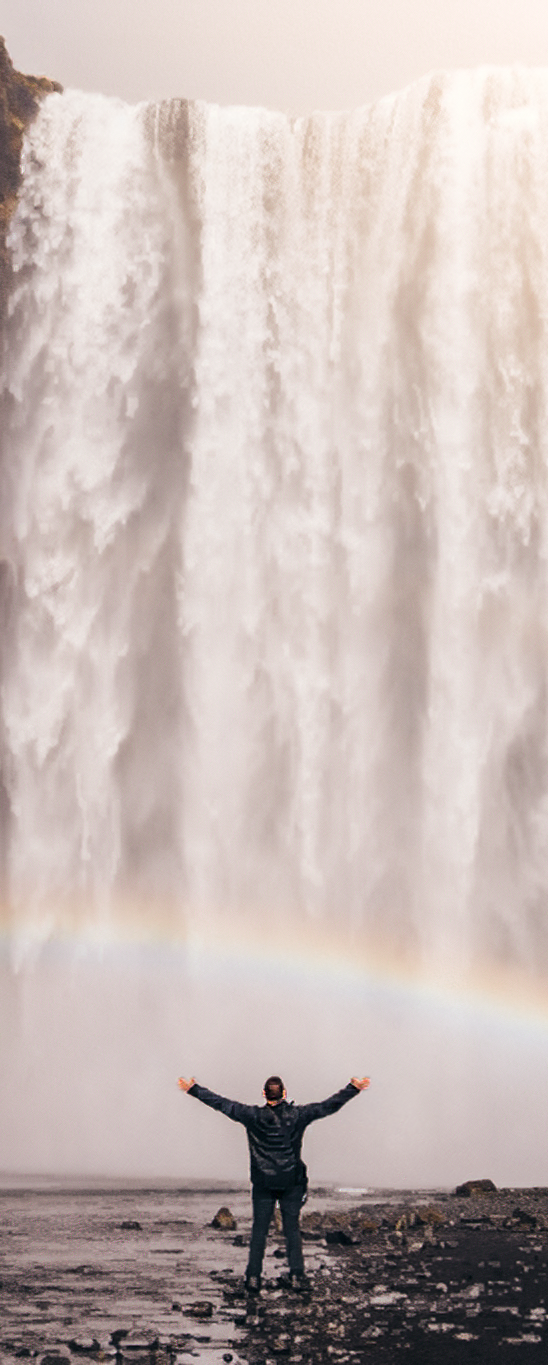}
\caption{Horizontal crop of image 16 from the CLIC2020 professional dataset (best viewed on screen).
\textbf{Far left:} Original image.
\textbf{Center left:} C3 optimized for MSE, compressed to 0.088 bits/pixel.
\textbf{Center right:} C3 optimized for LPIPS, compressed to 0.076 bits/pixel.
\textbf{Far right:} C3 optimized for WD with $\sigma=8$, compressed to 0.077 bits/pixel.
The background texture of the waterfall is largely flattened in the MSE version. The LPIPS version slightly improves on this, albeit at the cost of losing significant detail on the person standing in the foreground. Remarkably, even without employing a saliency model, WD preserves detail in the foreground better than LPIPS, while also reproducing the waterfall texture faithfully.
}
\end{figure*}

\begin{figure*}
\includegraphics[width=.333\linewidth]{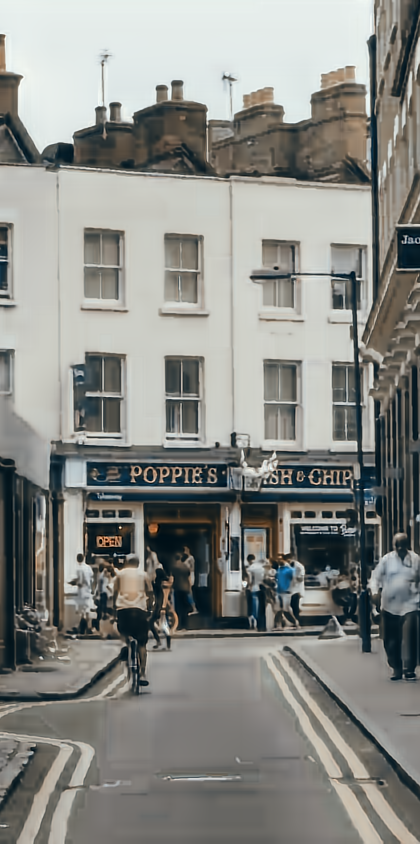}%
\includegraphics[width=.333\linewidth]{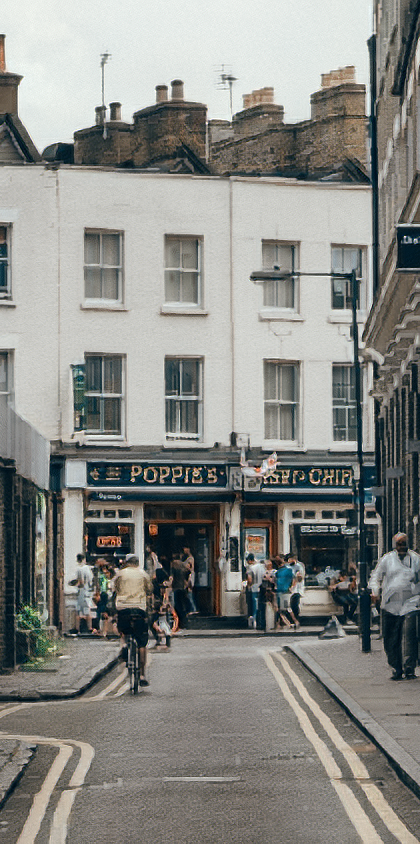}%
\includegraphics[width=.333\linewidth]{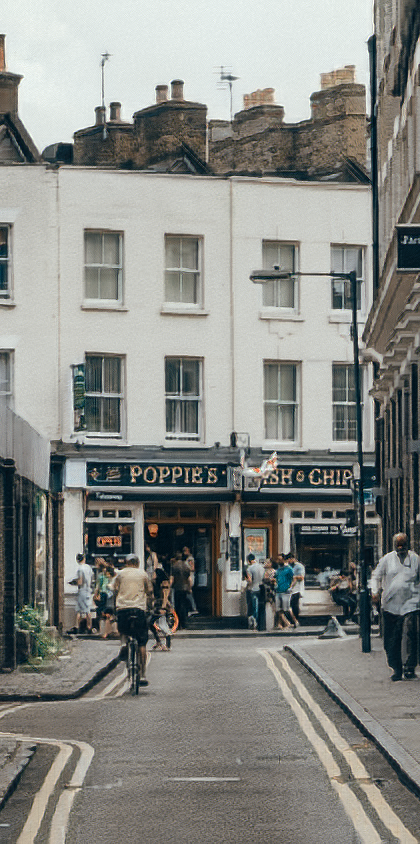}
\caption{Horizontal crop of image 11 from the CLIC2020 professional dataset (best viewed on screen).
\textbf{Left:} C3 optimized for MSE, compressed to 0.133 bits/pixel.
\textbf{Center:} C3 optimized for WD with $\sigma=8$, compressed to 0.134 bits/pixel.
\textbf{Right:} C3 optimized for WD with $\sigma$ derived from a saliency map, compressed to 0.130 bits/pixel.
Again, optimization for MSE leads to flattened texture, as seen in the reproduction of the street and chimneys, while the signage is reproduced well due to its high contrast (difference between dark and light), which MSE is sensitive to. While texture is vastly improved in the WD-optimized version, the flat $\sigma$ version struggles to reproduce the signage well. On the right, the saliency model assigns high saliency to that image region, leading to a better reconstruction and higher legibility of the text.}
\end{figure*}

\begin{figure*}
\includegraphics[width=.25\linewidth]{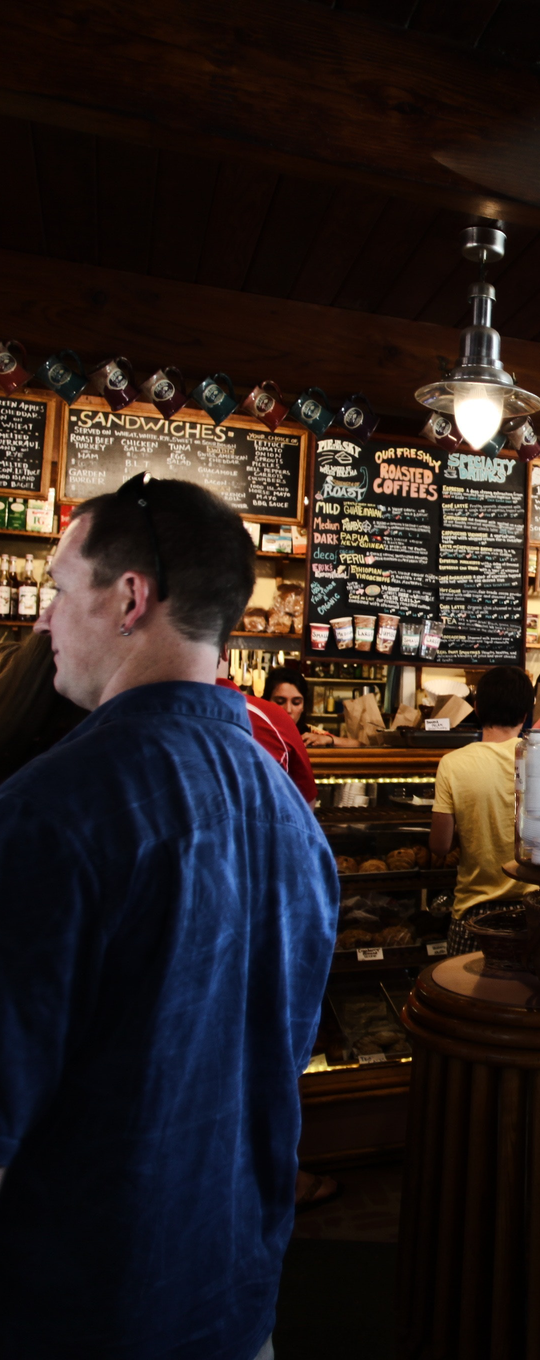}%
\includegraphics[width=.25\linewidth]{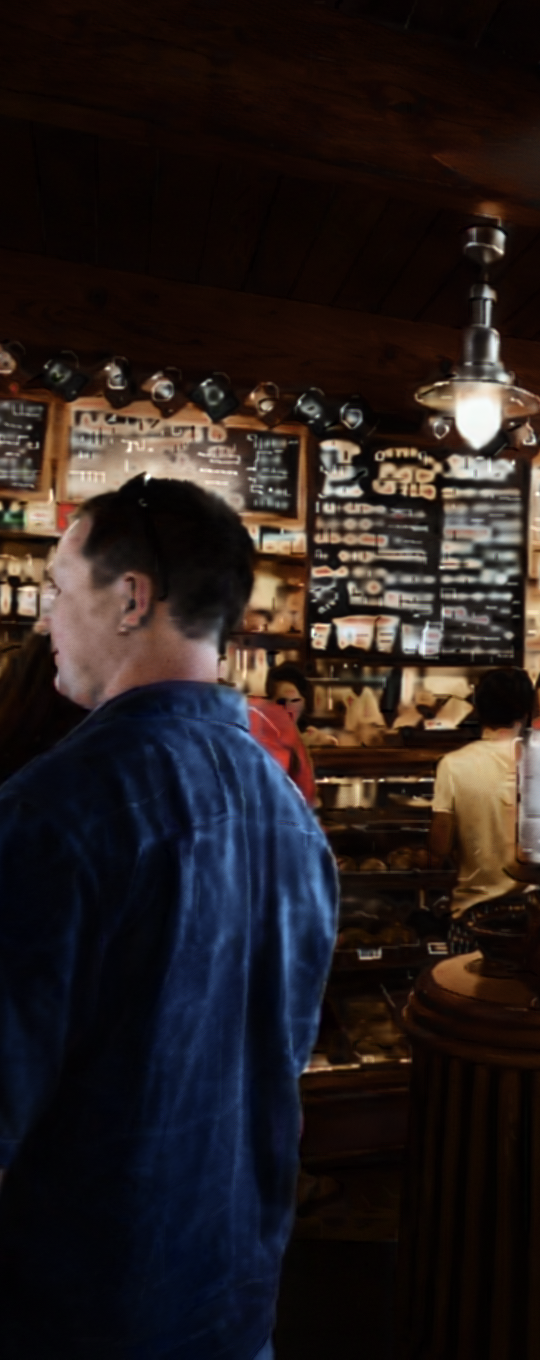}%
\includegraphics[width=.25\linewidth]{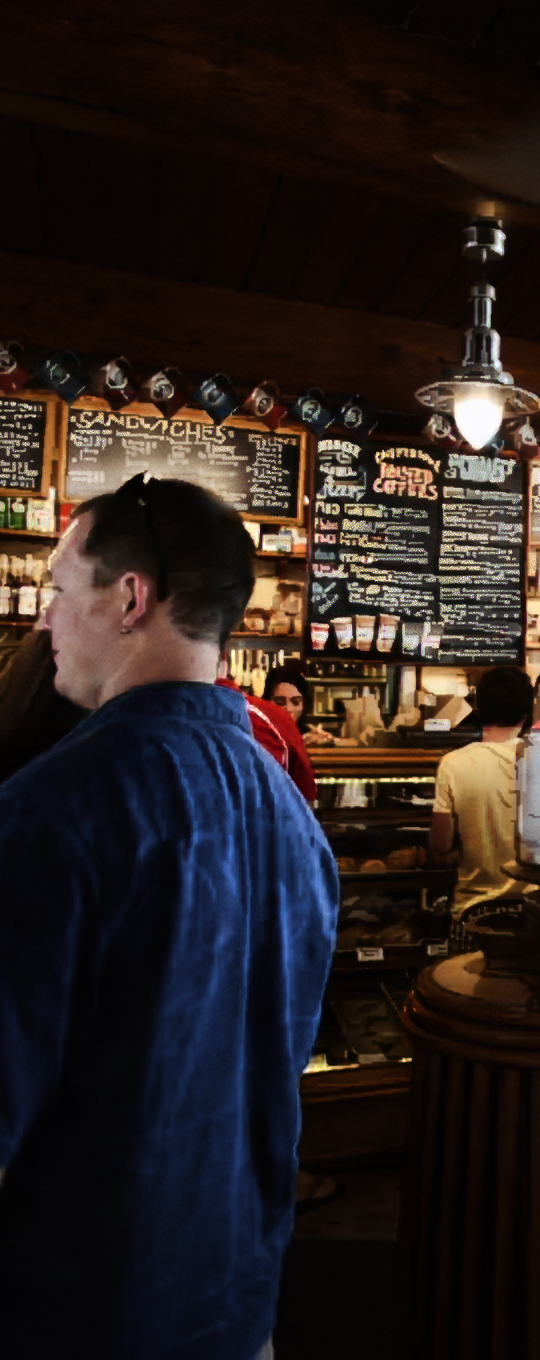}%
\includegraphics[width=.25\linewidth]{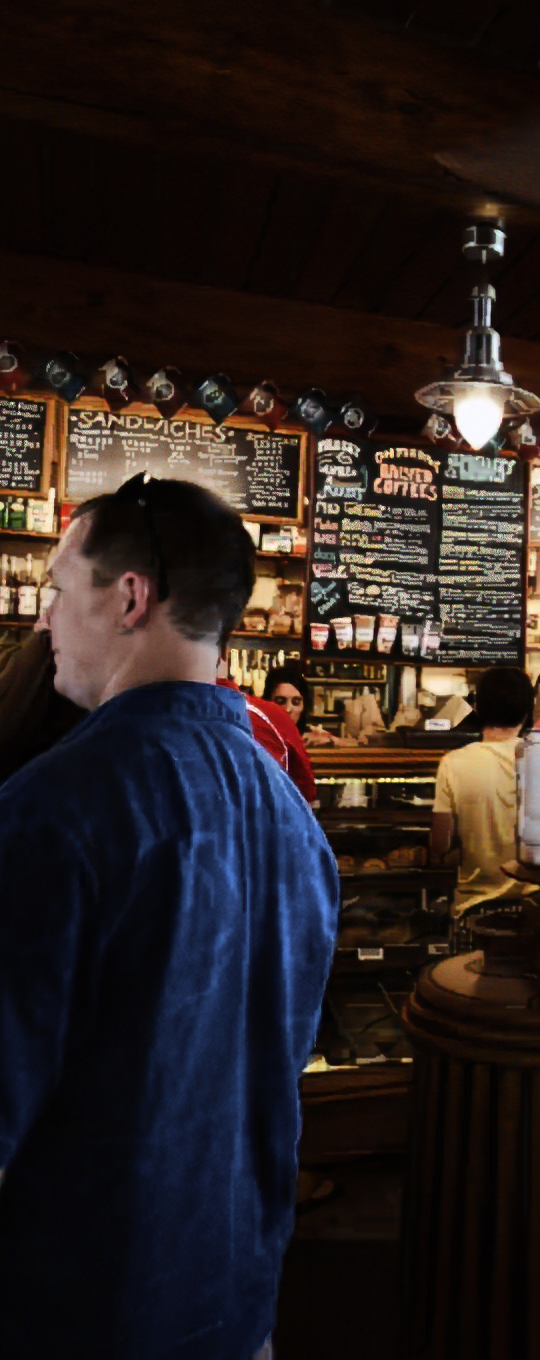}
\caption{Horizontal crop of image 13 from the CLIC2020 professional dataset (best viewed on screen).
\textbf{Far left:} Original image.
\textbf{Center left:} C3 optimized for LPIPS, compressed to 0.093 bits/pixel.
\textbf{Center right:} C3 optimized for WD with $\sigma=8$, compressed to 0.093 bits/pixel.
\textbf{Far right:} C3 optimized for WD with $\sigma$ derived from a saliency map, compressed to 0.091 bits/pixel.
An example of text in the background of the image reproduced as a visual texture. Note that text legibility doesn't improve when comparing the WD version with and without using a saliency model, as in this case, the text is not predicted to be a salient image region. However, the visual quality in both cases is still much preferable to the one achieved by optimizing for LPIPS.
}
\end{figure*}

\end{landscape}

\end{document}